\title{A Theory of Local Learning, the Learning Channel, and the Optimality of Backpropagation}
\author{
        \textsc{}
            \qquad
        \textsc{Pierre Baldi}\thanks{Contact author} \quad {and} \quad \textsc{ Peter Sadowski}
        \mbox{}\\ %
        Department of Computer Science\\
        University of California, Irvine\\
        Irvine, CA 92697-3435\\
        \mbox{}\\ %
        \normalsize
            \texttt{pfbaldi,psadowsk}
        \normalsize
            \texttt{@uci.edu}
}
\date{\today}
\documentclass[11pt]{article}

\usepackage[paper=a4paper,dvips,top=1.5cm,left=1.5cm,right=1.5cm,
    foot=1cm,bottom=1.5cm]{geometry}

\usepackage{times}
\usepackage{graphicx}
\usepackage[fleqn]{amsmath}
\usepackage{amsfonts}
\usepackage{amssymb}
\usepackage{amsthm}
\usepackage{amsopn}
\usepackage{xspace}
\usepackage{array}
\usepackage{epsfig}
\usepackage{tikz}
\usetikzlibrary{matrix}

\newcommand{\be}{\begin{equation}}
\newcommand{\ee}{\end{equation}}

\usepackage{url}
\usepackage{graphicx}
\usepackage{epsfig}
\usepackage{url}
\usepackage{textcomp}
\usepackage{booktabs}
\usepackage{pstricks}
\usepackage{setspace}
\usepackage[numbers]{natbib}

\usepackage{color}
\usepackage{amsfonts}
\usepackage{amsmath,amsthm,amscd}

\numberwithin{figure}{section}

\date{}
\begin{document}

\maketitle

\begin{abstract}
In a physical neural system, where storage and processing are intimately intertwined,  the rules for adjusting the synaptic weights can only depend on variables that are available locally, such as the activity of the pre- and post-synaptic neurons, resulting in {\it local learning rules}. A systematic framework for  studying the space of local learning rules is obtained by first specifying the nature of the local variables, and then the functional form that ties them together into each learning rule.
Such a framework enables also the systematic discovery of new learning rules 
and exploration of relationships between learning rules and group symmetries. We study polynomial local learning  rules stratified by their degree and analyze their behavior and capabilities in both linear and non-linear units and networks. 
Stacking local learning rules in deep feedforward networks leads to {\it deep local learning}. While deep local learning can learn interesting representations, it cannot learn complex input-output functions, even when targets are available for the top layer. Learning complex  input-output functions requires {\it local deep learning} where target information is communicated to the deep layers through a backward {\it learning channel}. The nature of the communicated information about the targets and the structure of the learning channel partition the space of learning algorithms. For any learning algorithm, the {\it capacity}  of the learning channel can be defined as the
number of bits provided about the error gradient per weight, divided by the number of required operations per weight. We estimate the capacity associated with several learning algorithms and show that backpropagation outperforms them by simultaneously maximizing the information rate and minimizing the computational cost. This result is also shown to be true for recurrent networks, by unfolding them in time. The theory clarifies the concept of Hebbian learning, establishes the power and limitations of local learning rules, introduces the learning channel which enables a formal analysis of the optimality of backpropagation, and explains the sparsity of the space of learning rules discovered so far.
\end{abstract}


\section{Introduction}
\label{sec:Introduction}

The deep learning problem can be viewed as the problem of learning the connection weights of a large computational graphs, in particular the weights of the deep connections that are far away from the inputs or outputs \cite{schmidhuber2015deep}. 
In spite of decades of research, 
only very few algorithms have been proposed to try to address this task. Among the most important ones, and somewhat in opposition to each other, are backpropagation \cite{rumelhart1986learning} and Hebbian learning \cite{hebb1949organization}. Backpropagation has been the dominant algorithm, at least in terms of successful applications, which have ranged over the years from computer vision
\cite{krizhevsky2012imagenet} to high-energy physics \cite{baldi2014searching}. In spite of many attempts, no better algorithm has been found, at least within the standard supervised learning framework.
In contrast to backpropagation which is a well defined algorithm--stochastic gradient descent--Hebbian learning has remained a more nebulous concept, often associated with notions of biological and unsupervised learning. While less successful than backpropagation in applications, it has periodically inspired the development of theories aimed at capturing the essence of neural learning \cite{hebb1949organization,fukushima1980neocognitron,hopfield1982}. Within this general context, the goal of this work is to create a precise framework to organize and study the space of learning rules and their properties and address several questions, in particular: (1) What is Hebbian learning? (2) What are the capabilities and limitations of Hebbian learning? (3) What are the connections between Hebbian learning and backpropagation? (4) Are there other learning algorithms better than backpropagation? These questions are addressed in two parts: the first part focuses on Hebbian learning, the second part on backpropagation.

\subsection{The Deep Learning Problem}

At the core of many neural system models is the idea that information is stored in synapses and typically represented by a ``synaptic weight''. While synapses could conceivably be far more complex (e.g.  \cite{NIPS2013_4872}) and require multiple variables for describing their states, for simplicity here we will use the single synaptic weight framework, although the same ideas can readily be extended to more complex cases. In this framework, synapses are faced with the task of adjusting their individual weights in order to store relevant information and collectively organize in order to sustain  neural activity leading to appropriately adapted behavior at the level of the organism. This is a daunting task  if one thinks about the scale of synapses and how remote they can be from sensory inputs and motor outputs. Suffice it to say that when rescaled by a factor of $10^6$, a synapse is the size of a fist and the bow of the violin, or the tennis racket, it ought to help control is 1,000 miles away. This is the core of the deep learning problem.

\subsection{The Hebbian Learning Problem}
Donald Hebb is credited with being among the first to think about this problem and attempt to come up with a plausible solution in his 1949 book {\it The Organization of Behavior}
\cite{hebb1949organization}.
However, Hebb was primarily a psychologist and his ideas were stated in rather vague terms, such as:
``When an axon of cell A is near enough to excite cell B and repeatedly or persistently takes part in firing it, some growth process or metabolic change takes place in one or both cells such that A's efficiency, as one of the cells firing B, is increased'' often paraphrased as ``Neurons that fire together wire together''. Not a single equation can be found in his book.

While the concept of Hebbian learning has played an important role in the development of both neuroscience and machine learning, its lack of crispness becomes obvious as soon as one raises simple questions like: Is the backpropagation learning rule Hebbian? Is Oja's learning rule \cite{oja1982simplified} Hebbian? Is a rule that depends on a function of the output \cite{hyvarinen1998independent}
Hebbian? Is a learning rule that depends only on the input Hebbian? and so forth. This lack of crispness is more than a simple semantic issue. While it may have helped the field in its early stages--in the same way that vague concepts like ``gene'' or ``consciousness'' may have helped molecular biology or neuroscience, it has also prevented clear thinking to address basic questions regarding, for instance, the behavior of linear networks under Hebbian learning, or the capabilities and limitations of Hebbian learning in both shallow and deep networks.

At the same time, there have been several attempts at putting 
the concept of Hebbian learning at the center of biological learning \cite{fukushima1980neocognitron,hopfield1982}.
Hopfield proposed to use Hebbian learning to store memories in networks of symmetrically connected threshold gates.
While the resulting model is elegant and amenable to interesting analyses, it oversimplifies the problem by considering only  shallow networks, where all the units are visible and have targets. Fukushima proposed the neocognitron architecture for computer vision, inspired by the earlier neurophysiological work of Hubel and Wiesel \cite{hubel1962receptive}, essentially in the form of a multi-layer convolutional neural network. Most importantly for the present work, Fukushima proposed to learn the parameters of the neocognitron architecture in a self-organized way using some kind of Hebbian mechanism. While the Fukushima program has remained a source of inspiration for several decades, a key result of this paper is to show that such a program {\it cannot} succeed at finding an optimal set of weights in a feedforward architecture, regardless of which specific form of Hebbian learning is being used. 

\subsection{The Space of Learning Rules and its Sparsity Problem}
 
Partly related to the nebulous nature of Hebbian learning, is the observation that so far the entire machine learning field has been able to come up only with very few learning rules like the backpropagation rule and Hebb's rule. Other familiar rules, such as the perceptron learning rule 
\cite{rosenblatt1958perceptron}, the delta learning rule \cite{widrow1960adaptive}, and Oja's rule \cite{oja1982simplified}, can be viewed as special cases of, or variations on, backpropagation or Hebb
(Table \ref{tab:space}). Additional variations are found also in, for instance,  
\cite{bienenstock1982theory,intrator1992objective,law1994formation},
and discussions of learning rules from a general standpoint in
\cite{baldi88b,kohonen1995self}.
This creates a potentially unsatisfactory situation given that of the two most important learning algorithms, the first one could have been derived by Newton or Leibniz, and the second one is shrouded in vagueness. Furthermore, this raises the broader question of the nature of the space of learning rules. In particular, why does the space seem so sparse? Are there new rules that remain to be discovered in this space? 

\begin{table}[h!]
\renewcommand{\arraystretch}{1.5}
\begin{center}
    \begin{tabular}{ | c | l |   }
    \hline
 \textbf{Learning Rule} &  \textbf{Expression}   \\ \hline
 Simple Hebb & $\Delta w_{ij} \propto O_iO_j$\\ \hline
  Oja &    $\Delta w_{ij} \propto O_iO_j-O_i^2w_{ij}$ \\ \hline
 Perceptron &$\Delta w_{ij} \propto   (T-O_i) O_j$ \\\hline
 Delta  & $\Delta w_{ij} \propto(T-O_i)f'(S_i)O_j$ \\ \hline
  Backprogation &  $\Delta w_{ij} \propto B_iO_j$ \\ \hline
    \end{tabular}
\end{center}
\caption{Common learning rules and their on-line expressions. 
$O_i$ represents the activity of the postsynaptic neuron, $O_j$ the activity of the presynaptic neuron, and $w_{ij}$ the synaptic strength of the corresponding connection.
$B_i$ represents the back-propagated error in the postsynaptic neuron. The perceptron and Delta learning rules were originally defined for a single unit (or single layer), in which case $T$ is the readily available output target.}
    \label{tab:space}
\end{table}

\section{A Framework for Local Learning Rules}
\label{sec:Framework}
The origin of the vagueness of the Hebbian learning idea is that 
it indiscriminately mixes two fundamental but distinct ideas: (1) learning ought to depend on local information associated with the pre- and post-synaptic neurons; and (2) learning ought to depend on the correlation between the activities of these neurons, yielding a spectrum of possibilities on how these correlations are computed and used to change the synaptic weights.
The concept of local learning rule, mentioned but not exploited in 
\cite{baldi88b}, is more fundamental than the concept of Hebbian learning rule, as it explicitly exposes the more general notion of locality, which is implicit but somehow hidden in the vagueness of the Hebbian concept. 

\subsection{The Concept of Locality}

To address all the above issues, the first observation is that in a physical implementation a learning rule to adjust a synaptic weight can only include {\it local} variables. Thus to bring clarity to the computational models, {\it one must first define which variables are to be considered local in a given model}.
Consider the backpropagation learning rule
$\Delta w_{ij} \propto B_iO_j$ where $B_i$ is the postsynaptic backpropagated error and $O_j$ is the presynaptic activity.
If the backpropagated error is not considered a local variable, then backpropagation is not a local learning rule, and thus is not Hebbian. If the backpropagated error is considered a local variable, then backpropagation may be Hebbian, both in the sense of being local and of being a simple product of local pre- and post-synaptic terms. 
[Note that, even if considered local, the backpropagated error may or may not be of the same nature (e.g. firing rate) as the presynaptic term, and this may invalidate its Hebbian character depending, again, on how one interprets the vague Hebbian concept.] 

Once one has decided which variables are to be considered local in a given model, then one can generally express a learning rule as 

\be 
\Delta w_{ij}=F({\rm local\,\,\, variables})
\label{eq:rule1}
\ee
for some function $F$. A systematic study of local learning requires a systematic analysis of many cases in terms of not only the functions $F$, but also in terms of the computing units and their transfer functions (e.g. linear, sigmoidal, threshold gates, rectified linear, stochastic, spiking), the network topologies (e.g. shallow/deep, autoencoders, feedforward/recurrent), and other possible parameters (e.g. on-line vs batch). 
Here for simplicity we first consider single processing units with input-output functions of the form

\be
O=f(S)=f(\sum_j w_jI_j)
\label{eq:single1}
\ee
where $I$ is the input vector and the transfer function $f$ is the identity in the linear case, or the [0,1] logistic function $\sigma_{[0,1]}(x)=1/(1+e^{-x})$, or the [-1,1] hyperbolic tangent function $\sigma_{[-1,1]}(x)=(1-e^{-2x})/(1+e^{-2x})$, or the corresponding threshold functions $\tau_{[0,1]}$ and 
$\tau_{[-1,1]}$. When necessary, the bias is included in this framework by considering that the input value $I_0$ is always set to 1 and the bias is provided by corresponding weight $w_0$.
In the case of a network of $N$ such units, we write 

\be
O_i=f(S_i)=f(\sum_j w_{ij}O_j)
\label{eq:}
\ee
where in general we assume that there are no self-connections 
($w_{ii}$=0). In general, the computing units can be subdivided into three subsets corresponding to input units, output units, and hidden units.
While this formalism includes both feedforward and recurrent networks, in the first part of the paper we will focus primarily on feedforward networks. However issues of feedback and recurrent networks will become important in the second part. 

Within this general formalism, we typically consider first that the local variables are the presynaptic activity, the postynaptic activity, and $w_{ij}$ so that 

\be
\Delta w_{ij}=F(O_i,O_j,w_{ij})
\label{eq:rule2}
\ee
In supervised learning, in a model where the target $T_i$ is  considered a local variable, the rule can have the more general form

\be
\Delta w_{ij}=F(T_i, O_i,O_j,w_{ij})
\label{eq:rule3}
\ee
For instance, we will consider cases where the output is clamped to the value $T_i$, or where the error signal $T_i-O_i$ is a component of the learning rule. The latter is the case in the perceptron learning algorithm, or in the deep targets algorithm described  below, with backpropagation as a special case. Equation \ref{eq:rule3} represents a local learning rule if one assumes that there is a target $T_i$ that is locally available. Targets can be clearly available and local for the output layer. However 
the generation and local availability of targets for deep layers is a fundamental, but separate, question that will be addressed in later sections. {\it Thus it is essential to note that the concept of locality is orthogonal to the concept of unsupervised learning}. An unsupervised learning rule can be non-local if $F$ depends on activities or synaptic weights that are far away in the network. Likewise a supervised learning rule can be local, if the target is assumed to be a local variable.
Finally, we also assume that the learning rate $\eta$ is a local variable contained in the function $F$. For simplicity, we will assume that the value of $\eta$ is shared by all the units, although more general models are possible.

In short, it is time to move away from the vagueness of the term ``Hebbian learning'' and replace it with a clear definition, in each situation, of: (1)  which variables are to be considered local; and (2) which functional form is used to combine the local variables into a local learning rule. A key goal is then
to systematically study the properties of different local rules across different network types.

\subsubsection{Spiking versus Non-Spiking Neurons}

The concept of locality (Equation 1) is completely general and applies equally well to networks of spiking neurons and non-spiking neurons. The analyses of specific local learning rules in Sections 3-5 are conducted for non-spiking neurons, but some extensions to spiking neurons are possible (see, for instance,
\cite{zenke2015diverse}). Most of the material in Sections 6-8 is general again and is applicable to networks of spiking units.  The main reason is that these sections are concerned primarily with the propagation of information about the targets from the output layer back to the deeper layers, regardless of how this information is encoded, and regardless of whether non-spiking or spiking neurons are used in the forward, or backward, directions.

\subsection{Coordinate Transformations and Symmetries}

This subsection is not essential to follow the rest of this paper and can initially be skipped. 
When studying local learning rules, it is important to look at the effects of coordinate transformations and various symmetries on the learning rules. While a complete treatment of these operations is beyond our scope, we give several specific examples below.   In general, applying coordinate changes or symmetries can bring to light some important properties of a learning rule, and shows in general that the function $F$ should not be considered too narrowly, but rather as a member of a class. 

\subsubsection{Example 1: Range Transformation (Affine Transformation)}

For instance, consider the narrow definition of Hebb's rule as 
$\Delta w_{ij}\propto O_iO_j$ applied to threshold gates with binary inputs.
This definition makes some sense if the threshold gates are defined using a $[-1,1]$ formalism, but is problematic over a $[0,1]$ formalism because 
it results in $\Delta w_{ij}=O_iO_j$ being 0 in three out of four cases, and always positive and equal to 1 in the remaining fourth case. Thus the narrow definition of Hebb's rule over a $[0,1]$ system should be modified using the corresponding affine transformation. However the new expression will have to be in the {\it same functional class}, i.e. in this case quadratic function over the activities. The same considerations apply when sigmoid transfer functions are used.

More specifically, [0,1] networks are transformed into [-1,1] networks through the transformation $x \to 2x-1$ or vice versa through the transformation $x \to (x+1)/2$. It is easy to show that a polynomial local rule in one type of network is transformed into a polynomial local rule of the same degree in the other type of network. For instance, a quadratic local rule with coefficients
$\alpha_{[0,1]}, \beta_{[0,1]}, \gamma_{[0,1]}, \delta_{[0,1]}$
of the form

\be
\Delta w_{ij} \propto \alpha_{[0,1]} O_iO_j  +\beta_{[0,1]} O_i +\gamma_{[0,1]} O_j +\delta_{[0,1]}
\label{eq:quadratic1}
\ee
is transformed into a rule with coefficients 
$\alpha_{[-1,1]}, \beta_{[-1,1]}, \gamma_{[-1,1]}, \delta_{[1,1]}$ through the homogeneous system:

\begin{eqnarray}
\alpha_{[-1,1]}&=&4\alpha_{[0,1]} \\
\beta_{[-1,1]}&=&2\beta_{[0,1]}-2\alpha_{[0,1]} \\
\gamma_{[-1,1]}&=&2\gamma_{[0,1]}-2\alpha_{[0,1]} \\
\delta_{[-1,1]}&=&\delta_{[0,1]}+\alpha_{[0,1]}-\beta_{[0,1]} -\gamma_{[0,1]}
\label{eq:transfo1}
\end{eqnarray}
Note that no non-zero quadratic rule can have the same form
in both systems, even when trivial multiplicative coefficients are absorbed into the learning rate. 

\subsubsection{Example 2: Permutation of Training Examples}
Learning rules may be more or less sensitive to permutations in the order in which examples are presented. In order to analyze the behavior of most rules, here we will assume that they are not sensitive to the order in which the examples are presented, which is generally the case if all the training examples are treated equally, and the on-line learning rate is small and changes slowly so that averages can be computed over entire epochs (see below).

\subsubsection{Example 3: Network Symmetries}

When the same learning rule is applied isotropically, it is important to examine its behavior under the symmetries of the network architecture to which it is applied.  This is the case, for instance, in Hopfield networks where all the units are connected symmetrically to each other (see next section), or between fully connected layers of a feedforward architecture. In particular, it is important to examine whether differences in inputs or in weight initializations can lead to symmetry breaking.
It is also possible to consider models where different neurons, or different connections, use different rules, or rules in the same class (like Equation 
\ref{eq:quadratic1}) but with different coefficients. 

\subsubsection{Example 4: Hypercube Isometries}
\begin{figure}[h!]
    \centering
    \includegraphics[width=0.8\textwidth]{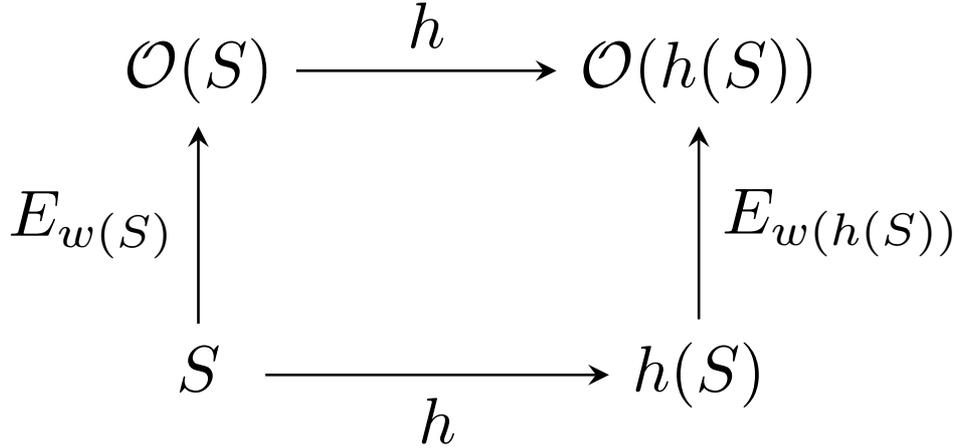}
    \caption{Commutative diagram for the simple Hebb rule in a Hopfield network. Application of the simple Hebb's rule to a set $S$ binary vectors over the $[-1,1]^N$ hypercube in a Hopfield network with $N$ units yields a set of symmetric weights $w_{ij}=w_{ji}$ and a corresponding quadratic energy function $E_{w(S)}$, which ultimately produces a directed acyclic orientation of the hypercube ${\cal O}(S)$ directing the dynamics of the network towards minima of $E_{w(S)}$. An isometry $h$ over the hypercube yields a new set of vectors $h(S)$ hence, by application of the same Hebb rule,  a new set of weights, a new energy function $E_{w(h(S))}$, and a new acyclic orientation such that $h({\cal O}(S))={\cal O}(h(S))$.}
    \label{fig:invariance}
\end{figure}

As a fourth example \cite{baldi87}
consider a Hopfield network \cite{hopfield1982} consisting of $N$ threshold gates, with $\pm1$ outputs, connected symmetrically to each other ($w_{ij}=w_{ji}$).
It is well known that such a system and its dynamics is characterized by the quadratic energy function $E=-(1/2)\sum_{i,j}w_{ij}O_iO_j$ (note that the linear terms of the quadratic energy function are taken into account by $O_0=1$). The quadratic function $E$ induces an acyclic orientation ${\cal O}$ of the 
$N$-dimensional Hypercube ${\cal H}^N=[-1,1]^N$ where the edge between two neighboring (i.e. at Hamming distance 1) state spaces $x$ and $y$ is oriented from $x$ to $y$ if and only if $E(x)>E(y)$.
Patterns or ``memories'' are stored in the weights of the system by applying the simple Hebb rule $\Delta w_{ij} \propto O_iO_j$ to the memories.
Thus a given training set $S$ produces a corresponding set of weights, thus a corresponding energy function, and thus a corresponding acyclic orientation ${\cal O} (S)$ of the hypercube.
Consider now an isometry $h$ of the $N$-dimensional hypercube, i.e. a one-to-one function from ${\cal H}^N$ to ${\cal H}^N$ that preserves the Hamming distance. It is easy to see that all isometries can be generated by composing two kinds of elementary operations: (1) permuting two components; and (2) inverting the sign of a component (hence the isometries are linear). It is then natural to ask what happens to 
 ${\cal O} (S)$ when $h$ is applied to ${\cal H}^N$ and thus to $S$.  It can be shown that under the simple Hebb rule the ``diagram commutes'' 
 (Figure \ref{fig:invariance}).
In other words, $h(S)$ is a new training set which leads to a new
acyclic orientation ${\cal O}(h(S))$ and 

\be
h({\cal O}(S))={\cal O}(h(S))
\label{eq:transfo2}
\ee
Thus the simple Hebb rule is invariant under the action of the isometries of the hypercube.
In Appendix A, we show it is the only rule with this property.

\subsection{Functional Forms}

Within the general assumption that $\Delta w_{ij}=F(O_i,O_j,w_{ij})$,
or $\Delta w_{ij}=F(T_i,O_i,O_j,w_{ij})$ in the supervised case,
one must consider next the functional form of $F$. Among other things, this allows one to organize and stratify the space of learning rules. As seen above, the function $F$ cannot be defined too narrowly, as it must be invariant to certain changes, and thus one is primarily interested in classes of functions.
In this paper, we focus exclusively on the case where $F$ is a polynomial function of degree $n$ (e.g. linear, quadratic, cubic) in the local variables, although other functional forms could be considered, such as rational functions or power functions with rational exponents. Most rules that are found in the neural network literature correspond to low degree polynomial rules.  Thus we consider functions $F$ comprising a sum of terms of the form $\alpha O_i^{n_i} O_j^{n_j}w_{ij}^{n_{w_{ij}}}$ (or 
$\alpha T_i^{n_{T_i}} O_i^{n_i} O_j^{n_j}w_{ij}^{n_{w_{ij}}}$)
where $\alpha$ is a real coefficient [in this paper we assume that the constant $\alpha$ is the same for all the weights, but many of the analyses carry over to the case where different weights have different coefficients although such a system is not invariant under relabeling of the neurons]; $n_{T_i}$, $n_i$, $n_j$, and $n_{w_{ij}}$ are non-negative integers satisfying $n_{T_i}+ n_i+n_j+n_{w_{ij}} \leq n$.
In this term, the {\it apparent} degree of $w$ is $n_{w_{ij}}$ but the {\it effective} degree of $w$ may be higher because $O_i$ depends also on $w_{ij}$, typically in a linear way at least around the current value of $O_i$. In this case, the effective degree of $w_{ij}$ in this term is $n_i+n_{w_{ij}}$.
[For instance, consider a rule of the form 
$\Delta w_{ij}=w_{ij}O_i^2I_j$, with a linear unit $O_i=\sum_k w_{ik}I_k$. The apparent degree of the rule in $w_{ij}$ is 1, but the effective degree is 3.]
Finally, we let $d$ ($d \leq n$) denote the highest effective degree of $w$, among all the terms in $F$. As we shall see, $n$ and $d$ are the two main numbers of interest used to stratify the polynomial learning rules.

\subsection{Terminology}
We do not expect to be able to change how the word Hebbian is used but the recommendation, used in the rest of this paper, is to replace Hebbian with the more precise concept of local learning rule, which assumes a pre-existing definition of which variables are to be considered local. Within the set of local learning rules, 
it is easy to see that in general linear ($n=1$) learning rules of the form $\Delta w_{ij} \propto \alpha O_i + \beta O_j + \gamma w_{ij}$ are not very useful (in fact the same is true also for rules of the form $\Delta w_{ij} \propto h(O_i) + g(O_j)+ k(w_{ij}) $ for any functions $h$, $g$, and $k$). Thus, for a local learning rule to be interesting it must be at least quadratic.

Within quadratic learning rules, one could adopt the position that only  $\Delta w_{ij} \propto O_iO_j$ should be called Hebbian. At the other extreme, one could adopt the position that all quadratic rules 
with $n=2$ should be called Hebbian.
This would include the correlation rule

\be 
\Delta w_{ij} \propto (O_i-E(O_i))(O_j-E(O_j))
\label{eq:}
\ee
which requires information about the averages 
$E(O_i)$ and $E(O_j)$ over the training examples, and other rules of the form

\be
\Delta w_{ij} \propto \alpha O_iO_j + \beta O_i + \gamma O_j + \delta
\label{eq:quadratic2}
\ee
Note that this is not the most general possible form since other terms can also be included (i.e. terms in $O_i^2$, $O_j^2$, $w_{ij}$, $w_{ij}^2$,
$w_{ij}O_i$, and $w_{ij}O_j$) and will be considered below.
Note also that under any of these definitions of Hebbian, Oja's rule

\be
\Delta w_{ij} \propto O_iO_j - O_i^2 w_{ij}
\label{eq:}
\ee
is local, but not Hebbian since it is not quadratic in the local variables $O_i$, $O_j$, and $w_{ij}$. Rather it is a cubic rule with $n=3$ and $d=3$. 

In any case, to avoid these terminological complexities, which result from the vagueness of the Hebbian concept, we will: (1) focus on the concept of locality; (2) stratify local rules by their degrees $n$ and $d$; and (3) reserve ``simple Hebb'' for the rule  $\Delta w_{ij} \propto O_iO_j$, avoiding to use ``Hebb'' in any other context.  

\subsection{Time-scales and Averaging}

We assume a training set consisting of $M$ inputs $I(t)$ for $t=1, \ldots, M$ in the unsupervised case, and $M$ input-target pairs 
$(I(t),T(t))$ in the supervised case.
On-line learning with local rules will exhibit stochastic fluctuations and the weights will change at each on-line presentation. However, with a small learning rate and randomized order of example presentation, we expect the long term behavior to be dominated by the average values of the weight changes computed over one epoch. Thus we assume that
$O$ varies rapidly over the training data, compared to the synaptic weights $w$ which are assumed to remain constant over an epoch.
The difference in time-scales is what enables the analysis since we assume the weight $w_{ij}$ remains essentially constant throughout an epoch and we can compute the average of the changes induced by the training data over an entire epoch.
While the instantaneous evolution of the weights is governed by the relationship

\be
w_{ij}(t+1)=w_{ij} (t) + \eta \Delta w_{ij} (t)
\label{eq:evolution1}
\ee
the assumption of small $\eta$ allow us to average this relation over an entire epoch and write

\be
w_{ij}(k+1)=w_{ij} (k) + \eta E(\Delta w_{ij})
\label{eq:evolution2}
\ee
where the index $k$ is now over entire epochs, and $E$ is the expectation taken over the corresponding epoch. Thus, in the analyses, we must first compute the expectation $E$ and then solve the recurrence relation (Equation \ref{eq:evolution2}), or the corresponding differential equation.

\subsection{Initial Roadmap}

The article is subdivided into two main parts. In the first part, the focus is on Hebbian learning, or more precisely on local learning rules. Because we are restricting ourselves to learning rules with a polynomial form, the initial goal is to estimate expectations of the form
$E( O_i^{n_i} O_j^{n_j}w_{ij}^{n_{w_{ij}}})$ in the unsupervised case, or $E(T_i^{n_{T_i}} O_i^{n_i} O_j^{n_j}w_{ij}^{n_{w_{ij}}})$ in the supervised case. Because of the time-scale assumption, within an epoch we can assume that $w_{ij}$ is constant and therefore the corresponding term factors out of the expectation. Thus we are left with estimating terms of the form
$E( O_i^{n_i} O_j^{n_j})$ in the unsupervised case, or $E(T_i^{n_{T_i}} O_i^{n_i} O_j^{n_j})$ in the supervised case. 

In terms of architectures, we are primarily interested in deep feedforward architectures and thus we focus first on layered feedforward networks, with local supervised or unsupervised learning rules, where local learning is applied layer by layer in batch mode, starting from the layer closest to the inputs.
{\it In this feedforward framework, within any single layer of units, all the units learn independently of each other given the inputs provided by the previous layer. Thus in essence the entire problem reduces to understanding learning in a single unit and,
using the notation of Equation \ref{eq:single1}, to estimating the expectations
$E( O^{n_O} I_j^{n_j})$ in the unsupervised case, or $E(T^{n_{T}} O^{n_O} I_j^{n_j})$ in the supervised case, where $I_j$ are the inputs and $O$ is the output of the unit being considered.} In what follows, we first consider the linear case (Section \ref{sec:Linear}) and then the non-linear case (Section \ref{sec:NonLinear}). We then 
give examples of how new learning rules can be derived (Section \ref{sec:NewRules}).

In the second part, the focus is on backpropagation. We first 
study the limitations of purely local learning in shallow or deep networks, also called deep local learning (Section \ref{sec:Limitations}). 
To go beyond these limitations, naturally leads to the introduction of local deep learning algorithms and deep targets algorithms, and the study of the properties of the backward learning channel and the optimality of backpropagation (Sections 7 and 8).

\section{Local Learning in the Linear Case}
\label{sec:Linear}
The study of feedforward layered linear networks is thus reduced to the study of a single linear unit of the form $O=\sum_{i=0}^N w_i I_i$. In this case, to understand the behavior of any local learning rule,  one must compute expectations of the form

\be
E(T^{n_T}O^{n_O}I_i^{n_{I_i}}w_{i}^{n_{i}})=
w_{i}^{n_{i}}E \left [T^{n_T}(\sum_k w_k I_k)^{n_O}I_i^{n_{I_i}}
\right ]
\label{eq:}
\ee
This encompasses also the unsupervised case by letting $n_T=0$.
Thus this expectation is a polynomial in the weights, with coefficients that correspond to the statistical moments of the training data of the form $E(T^{n_T}I_i^{n_\alpha}I_k^{n_\beta})$. When this polynomial is linear in the weights ($d \leq 1$), the learning equation can be solved exactly using standard methods. When  the effective degree is greater than 1 ($d>1$),
then the learning equation can be solved in some special cases, but not in the general case.

To look at this analysis more precisely, here we assume that the learning rule only uses data terms of order two or less. Thus only the means, variances, and covariances of $I$ and $T$ are necessary to compute the expectations in the learning rule. For example, a term $w_iTO$ is acceptable, but not
$w_iTO^2$ which requires third-order moments of the data of the form $E(TI_iI_j)$ to compute its expectation. 
To compute all the necessary expectations systematically, we will use the following notations.

\subsection{Notations}

\begin{itemize}
\item All vectors are column vectors.
\item $A'$ denotes the transpose of the matrix $A$, and similarly for vectors.
\item $u$ is the $N$ dimensional vector of all ones: $u'=(1,1, \ldots,1)$.
\item $\circ$ is the Hadamard or Schur product, i.e. the component-wise product of matrices or vectors of the same size. We denote by $v^{(k)}$
the Schur product of $v$ with itself $k$ times, i.e. $v^{(k)}=v \circ v \ldots \circ v$.
\item 
${\rm diag} M$ is an operator that creates a vector whose components are the diagonal entries of the square matrix $M$.
\item When applied to a vector ${\rm Diag} v$  represents the 
square diagonal matrix whose components are the 
components of the vector $M$. 
\item  ${\rm Diag} M$ represents the square diagonal matrix whose entries on the diagonal are identical to those of $M$ (and 0 elsewhere), when $M$ is a square matrix.
\item For the first order moments, we let $E(I_i)=\mu_i$ and $E(T)=\mu_T$. In vector form, $\mu=(E(I_i))$.
\item For the second order moments, we introduce the matrix
$\Sigma_{II'}=(E(I_iI_j))=({\rm Cov}(I_i,I_j)+\mu_i\mu_j)$
\end{itemize}

\subsection{Computation of the Expectations}
With these notations, we can compute all the necessary expectations.
Thus:
\begin{itemize}
\item In Table \ref{tab:linearterms}, we list all the possible terms with $n=0$ or
$n=1$ and their expectations.
\item In Table \ref{tab:quadraticterms},  we list all the possible quadratic terms with $n=2$ and their expectations.
\item In Table \ref{tab:cubicterms},  we list all the possible cubic terms with $n=3$, requiring only first and second moments of the data, and their expectations.
\item In Table \ref{tab:ordernterms},  we list all the possible terms of order $n$,  requiring only first and second moments of the data, and their expectations.
\end{itemize}

Note that in Table \ref{tab:ordernterms},
for the term $w_i^{n-2}I_i^2$ the expectation in matrix form can be written as $w^{(n-2)} \circ {\rm Diag} \Sigma_{II'}={\rm Diag}(\Sigma_{II'})w \circ w^{(n-3)}$. Thus in the cubic case where $n=3$, the expectation has the form ${\rm Diag}(\Sigma_{II'})w$. Likewise, for  $w_i^{n-2}I_iT$
the expectation in matrix form can also be written as
$w^{(n-2)}\circ \Sigma_{IT'}=w^{(n-3)} \circ ({\rm diag }\Sigma_{IT'}) w$. Thus in the cubic case where $n=3$, the expectation has the form  $({\rm diag }\Sigma_{IT'}) w$.

Note also that when there is a bias term, we consider that the corresponding input $I_0$ is constant and clamped to 1, so that 
$E(I_0^n)=1$ for any $n$, and $I_0$ can simply be ignored in any product expression.
  
\subsection{Solving the Learning Recurrence Relation in the Linear Case ($d\leq 1$)}

When the effective degree $d$ satisfies $d \leq 1$, then the recurrence relation provided by Equation \ref{eq:recurrence0} is linear for any 
value of the overall degree $n$. Thus it can be solved by standard methods provided all the necessary data statistics are available to compute the expectations. More precisely, computing the expectation over one epoch leads to the relation

\be
w(k+1)=Aw(k)+ b
\label{eq:recurrence0}
\ee
Starting from $w(0)$ and iterating this relation, the solution can be written as
\be
w(k)=A^kw(0)+A^{k-1}b+A^{k-2} b + \ldots +Ab +b=A^kw(0)+[{\rm I}+A+A^2+\ldots+A^{k-1}]b
\label{eq:recurrence1}
\ee
where $\rm I$ denotes the identity matrix. Furthermore, if $A-{\rm I}$ is an invertible matrix, this expression can be written as
 
 \be
w(k)=A^kw(0)+ (A^k-{\rm I})(A-{\rm I})^{-1}b=A^kw(0)+ (A-{\rm I})^{-1}(A^k-{\rm I})b
\label{eq:recurrence2}
\ee
When $A$ is symmetric, there is an orthonormal matrix $C$ such that $A=CDC^{-1}=CDC'$, where $D={\rm Diag}(\lambda_1,\ldots,\lambda_N)$ is a diagonal matrix and $\lambda_1, \ldots,\lambda_N$ are the real eigenvalues of $A$. Then for any power $k$ we have $A^k=CD^kC^{-1}=C{\rm Diag}(\lambda_1^k,\ldots,\lambda_N^k)C^{-1}$ and $A^k-{\rm I}=C (D^k-{\rm I}) C^{-1}=
C {\rm diag}(\lambda_1^k-1,\ldots,\lambda_N^k-1) C^{-1}$ so that
Equation \ref{eq:recurrence1} becomes

\be
w(k)=CD^kC^{-1}w(0)+ C[{\rm I}+D+D^2+\ldots+D^{k-1}]C^{-1}b=
CD^k C^{-1}w(0)+ CEC^{-1}b
\label{eq:recurrence3}
\ee
where $E={\rm Diag}(\xi_1,\ldots,\xi_N)$ is a diagonal matrix with $\xi_i=(\lambda_i^n-1)/(\lambda_i-1)$ if $\lambda_i \not = 1$, and $\xi_i=k$ if $\lambda_i = 1$. If all the eigenvalues of $A$ are between 0 and 1 ($0<\lambda_i<1$ for every $i$) then the vector $w(k)$ converges to the vector
$C \rm{Diag}(1/(1-\lambda_1),\ldots,1/(1-\lambda_N))C'b$.
If all the eigenvalues of $A$ are 1, then $w(k)=w(0)+kb$.

\begin{table}[h!]
\renewcommand{\arraystretch}{1.5}
\begin{center}
    \begin{tabular}{ | c | c | c |  }
    \hline
 \textbf{Constant and Linear Terms} &  \textbf{Expectation} & \textbf{Matrix Form} \\ \hline
  $c_i$ $(0,0)$ &$c_i$ &$c=(c_i)$ \\ \hline
   $I_i$ $(1,0)$ &$\mu_i$ &$ \mu=(\mu_i)$ \\ \hline
    $O $ $(1,1)$& $\sum_j w_j\mu_j $&$ (w'\mu)u=(\mu ' w)u=({\rm Diag}\mu) w$ \\\hline
  $w_i$ $(1,1)$ &$w_i$&$w=(w_i)$ \\ \hline
     \hline
  $T$ $(1,0)$  & $\mu_T$ & $\mu_T u$\\\hline
    \end{tabular}
\end{center}
\caption{Constant and Linear Terms and their Expectations in Scalar and Vector Form. The table contains all the constant and linear terms of degree $(n,d)$ equal to $(0,0)$, $(1,0)$, and $(1,1)$ depending only on first order statistics of the data. 
The horizontal double line separates unsupervised terms (top) from supervised terms (bottom). The terms are sorted by increasing values of the effective degree ($d$), and then by increasing values of the apparent degree of $w_i$.}
    \label{tab:linearterms}
\end{table}

\begin{table}[h!]
\renewcommand{\arraystretch}{1.5}
\begin{center}
    \begin{tabular}{ | c | c | c |  }
    \hline
    \textbf{Quadratic Terms} &  \textbf{Expectation} & \textbf{Vector Form} \\ \hline
     $I_i^2$ $(2,0)$& ${\rm Var}I_i+\mu_i^2$ &${\rm diag}(\Sigma_{II'}) $\\\hline
  $I_iO$ $(2,1)$ &$w_i({\rm Var}I_i+\mu_i^2)+\sum_{j\not = i} w_j({\rm Cov}(I_i,I_j)+\mu_i\mu_j)$ &$({\rm Cov}I) w+ (\mu\mu')w=\Sigma_{II'}w$ \\ \hline      
$w_iI_i$ $(2,1)$&$w_i\mu_i$ &$w \circ \mu=({\rm Diag} \mu) w$ \\ \hline
   $O^2$ $(2,2)$ &$\sum_i w_i^2({\rm Var}I_i+\mu_i^2)+\sum_{i<j}2w_iw_j({\rm Cov}(I_i,I_j)+\mu_i\mu_j)$&$(w'\Sigma_{II'}w )u$ \\\hline
   $w_iO$ $(2,2)$ &$w_i\sum_jw_j\mu_j $&$(w'\mu)w=(\mu'w)w$ \\ \hline
    $w_i^2 $ $(2,2)$& $w_i^2$ &$w^{(2)}=(w_i^2)=w\circ w$ \\\hline 
  \hline
  $I_iT$ $(2,0)$  &${\rm Cov}(I_i,T)+\mu_i\mu_T$&${\rm Cov}(I,T)+\mu_T \mu=\Sigma_{IT'}$ \\\hline
  $T^2$ $(2,0)$ &${\rm Var}T+\mu_T^2$&$({\rm Var}T+\mu_T^2)u$ \\\hline
   $OT$ $(2,1)$ &$\sum_iw_i [{\rm Cov}(I_i,T)+\mu_i\mu_T]$&$w' \Sigma_{IT'}$ \\\hline
     $w_iT$ $(2,1)$ &$w_i \mu_T$&$\mu_T w $ \\\hline
        \end{tabular}
\end{center}
\caption{Quadratic Terms and their Expectations in Scalar and Vector Form. The table contains all the quadratic terms $(n,d)$ where $n=2$ and $d=0,1,$ or $2$. These terms depend only on the first and second order statistics of the data. The horizontal double line separates unsupervised terms (top) from supervised terms (bottom). The terms are sorted by increasing values of the effective degree ($d$), and then by increasing values of the apparent degree of $w_i$.}
    \label{tab:quadraticterms}
\end{table}

\begin{table}[h!]
\renewcommand{\arraystretch}{1.5}
\begin{center}
    \begin{tabular}{ | c | c | c |  }
    \hline
    \textbf{Simple Cubic Terms} &  \textbf{Expectation} & \textbf{Vector Form} \\ \hline
      $w_iI_i^2$ $(3,1)$&$w_i({\rm Var} I_i+\mu_i^2)$ &$w \circ {\rm diag} \Sigma_{II'}={\rm Diag}(\Sigma_{II'})w $
\\ \hline
   $w_iI_iO$ $(3,2)$ &$w_i[w_i({\rm Var}I_i+\mu_i^2)+\sum_{j\not = i} w_j({\rm Cov}(I_i,I_j)+\mu_i\mu_j)]$ &$w\circ \Sigma_{II'}w
      $  \\ \hline
            $w_i^2I_i$ $(3,2)$ &$w_i^2\mu_i$ &$w^{(2)} \circ \mu= w \circ
      ({\rm Diag} \mu) w$ \\ \hline
          $w_iO^2$ $(3,3)$ &$w_i[\sum_i w_i({\rm Var}I_i+\mu_i^2)+\sum_{i<j}2w_iw_j({\rm Cov}(I_i,I_j)+\mu_i\mu_j)]$ &
    $(w'\Sigma_{II'}w) w $ \\ \hline
       $w_i^2O$ $(3,3)$ &$w_i^2 \sum_j w_j\mu_j$ &$({\rm Diag}\mu w)\circ w^{(2)} $ \\ \hline
       $w_i^3$ $(3,3)$ &$w_i^3$ &$w^{(3)}=(w_i^3)=w \circ w \circ w$ \\ \hline\hline
     $w_iI_iT$ $(3,1)$ & $w_i({\rm Cov}(I_i,T)+\mu_i\mu_T)$ & $w\circ \Sigma_{IT'}={\rm diag }\Sigma_{IT'} w$\\\hline 
    $w_iT^2$ $(3,1)$ & $w_i({\rm Var} T + \mu_T^2)$ & $({\rm Var} T + \mu_T^2)w$\\\hline  
     $w_iOT$ $(3,2)$ & $w_i^2E(I_iT)+\sum_{j\not = i} w_iw_jE(I_jT)$ & $w \circ (w' \Sigma_{IT'})$\\\hline  
   $w_i^2T$ $(3,2)$ & $w_i^2\mu_T$ & $\mu_T w^{(2)}=\mu_T w \circ w$\\\hline
      \end{tabular}
\end{center}
\caption{Cubic Terms and their Expectations in Scalar and Vector From. The table contains all the terms of degree $(n,r)$ with $n=3$ and $r=0,1,2$ or $3$ that depend only on the first and second order statistics of the data. The horizontal double line separates unsupervised terms (top) from supervised terms (bottom). The terms are sorted by increasing values of the effective degree ($d$), and then by increasing values of the apparent degree of $w_i$.}
    \label{tab:cubicterms}
\end{table}

\begin{table}[h!]
\renewcommand{\arraystretch}{1.5}
\begin{center}
    \begin{tabular}{ | c | c | c |  }
    \hline
    \textbf{Simple}  &  \textbf{Expectation} & \textbf{Vector Form} \\ 
    \textbf{ $n$-th Terms} &  & \textbf{}\\    
    \hline
  $w_i^{n-2}I_i^2$ $(n,n-2)$ &$w_i^{n-2}({\rm Var} I_i+\mu_i^2)$ &$w^{(n-2)} \circ {\rm diag} \Sigma_{II'}$ \\ \hline
     $w_i^{n-2}I_iO$ $(n,n-1)$&$w_i^{n-2}[w_i({\rm Var}I_i+\mu_i^2)+\sum_{j\not = i} w_j({\rm Cov}(I_i,I_j)+\mu_i\mu_j)]$ &
   $w^{(n-2)}\circ \Sigma_{II'}w
      $ \\ \hline
       $w_i^{n-1}I_i$ $(n,n-1)$&$w_i^{n-1} \mu_i$ &$w^{(n-1)} \circ \mu=w^{n-2} \circ  ({\rm Diag} \mu)  w$\\ \hline
        $w_i^{n-2}O^2$ $(n,n)$&$w_i^{n-2}[\sum_i w_i({\rm Var}I_i+\mu_i^2)+\sum_{i<j}2w_iw_j({\rm Cov}(I_i,I_j)+\mu_i\mu_j)]$ &
    $    w' \Sigma_{II'}w w^{(n-2)} $ \\ \hline
  $w_i^{n-1}O$ $(n,n)$&$w_i^{n-1} \sum_j w_j\mu_j$ &$ w^{(n-1)}
       \circ ({\rm Diag}\mu) w $ \\ \hline
        $w_i^n$ $(n,n)$&$w_i^n$ &$w^{(n)}=(w_i^n)=w \circ \ldots \circ w$ \\ \hline\hline
          $w_i^{n-2}I_iT$ $(n,n-2)$& $w_i^{(n-2)}({\rm Cov}(I_i,T)+\mu_i\mu_T)$ & $w^{(n-2)}\circ \Sigma_{IT'}$\\\hline
$w_i^{n-2}T^2$ $(n,n-2)$& $w_i^{n-2}({\rm Var} T + \mu_T^2)$ & $
 ({\rm Var} T + \mu_T^2)w^{(n-2)}$\\\hline
      $w_i^{n-2}OT$ $(n,n-1)$& $w_i^{n-1}E(I_iT)+\sum_{j\not = i} w_i^{n-2}w_jE(I_jT)$ & $w^{(n-2)} \circ ( w' \Sigma_{IT'})$\\\hline 
   $w_i^{n-1}T$ $(n,n-1)$& $w_i^{n-1}\mu_T$ & $\mu_T w^{(n-1)}=\mu_T w \circ \ldots \circ w$\\\hline
         \end{tabular}
\end{center}
\caption{Simple Terms of Order $n$ and their Expectations in Scalar and Vector Form. The table contains all the terms of degree $(n,d)$ with $d=n-2, n-1$, or $n$  that depend only on the first and second order statistics of the data. The horizontal double line separates unsupervised terms (top) from supervised terms (bottom). The terms are sorted by increasing values of the effective degree ($d$), and then by increasing values of the apparent degree.}
    \label{tab:ordernterms}
\end{table}

\subsection{Examples}
We now give a few examples of learning equations with $d \leq 1$.

\subsubsection{Unsupervised Simple Hebbian Rule}
As an example, consider the simple Hebb rule with 
$\Delta w_i=\eta I_iO$ ($n=2$, $d=1$). Using Table \ref{tab:quadraticterms} we get in vector form 
$E(\Delta w)=\eta \Sigma_{II'}w$ and thus 

\be
w(k)=({\rm I}+\eta \Sigma_{II'})^k w(0)
\label{eq:}
\ee
In general, this will lead to weights that grow in magnitude exponentially with the number of epochs. For instance, if all the inputs have mean 0 ($\mu=0$), variance $\sigma_i^2$, and are independent of each other, then

\be 
w_i(k)=(1+\eta \sigma^2_i)^k w_i(0)
\label{eq:}
\ee
Alternatively, we can use the independence approximation to write

\be
w_i(k) =w_i(k-1) + \eta E(O(k-1)I_i) \approx  w_i(k-1) + \eta \mu_i E(O(k-1))=
w_i(k-1) + \eta \mu_i \sum_j w_j(k-1)\mu_j
\label{eq:}
\ee
which, in vector form, gives the approximation

\be
w(k) = w(k-1) +\eta \mu' w(k-1) \mu =({\rm I} + \eta A) w(k-1) \quad {\rm or} \quad w(k)= ({\rm I} +\eta A)^k w(0)
\label{eq:}
\ee
where $A=\mu \mu'$.

\subsubsection{Supervised Simple Hebbian Rule (Clamped Case)}
As a second example, consider the supervised version of the simple Hebb rule where the output is clamped to some target value $T$
with $\Delta w_i=\eta I_iT$ ($n=2$, $d=0$).
Using Table \ref{tab:quadraticterms} we get in vector form 
$E(\Delta w)=\eta \Sigma_{IT'}$ and thus 

\be
w(k)=w(0) +\eta k \Sigma_{IT'}
\label{eq:}
\ee
In general the weights will grow in magnitude linearly with the number $k$ of epochs, unless $E(I_iT)=0$ in which case the corresponding weight remains constant $w_i(k)=w_i(0)$.

Note that in some cases it is possible to assume, as a quick approximation, that the targets are independent of the inputs so that $E(\Delta w_i)=\eta E(T)E(I_i)=\eta E(T)\mu_i$. This simple 
approximation gives

\be
w(k)=w(0) +\eta k E(T) \mu
\label{eq:}
\ee
Thus the weights are growing linearly in the direction of the center of gravity of the input data.

Thus, in the linear setting, many local learning rules lead to divergent weights. There are notable exceptions, however, in particular when the learning rule is performing some form of (stochastic) gradient descent on a convex objective function.

\subsubsection{Simple Anti-Hebbian Rule}

The anti-Hebbian quadratic learning rule 
$\Delta w_i=-\eta I_iO$ ($n=2$, $d=1$) performs gradient descent on the objective function ${1\over 2}\sum_t O^2(t)$ and will tend to converge to the uninteresting solution where all weights (bias included) are equal to zero. 

\subsubsection{Gradient Descent Rule}
A more interesting example is provided by the rule
$\Delta w_i = \eta (T-O)I_i$ ($n=2$, $d=1$). Using Table \ref{tab:quadraticterms} we get in vector form 
$E(\Delta w)=\eta(\Sigma_{IT'}-\Sigma_{II'})$. The rule is convergent (with properly decreasing learning rate $\eta$) because it performs gradient descent on the quadratic error function 
${1 \over 2} \sum_{t=1}^M (T(t)-O(t))^2$ converging in general to the linear regression solution.

\par\null
\par

In summary, when $d \leq 1$ the dynamics of the learning rule can be solved exactly in the linear case and it is entirely determined by the statistical moments of the data, in particular by the means, variances, and covariances of the inputs and targets (e.g. when $n \leq 2$).

\subsection{The Case $d \geq 2$}

When the effective degree of the weights is greater than one in the learning rule, the recurrence relation is not linear and there is no systematic solution in the general case.  It must be noted however that in some special cases, this can result in a Bernoulli or Riccati ($d=2$) differential equation for the evolution of  each weight which can be solved (e.g. \cite{polyanin2002solutions}). 
For reasons that will become clear in later sections, let us for instance consider the learning equation

\be
\Delta w_i=\eta (1-w_i^2)I_i
\label{eq:}
\ee
with $n=3$ and $d=2$.
We have

\be
E(\Delta w_i)=\eta (1-w_i^2)\mu_i
\label{eq:}
\ee
Dropping the index $i$, the corresponding Riccati differential equation is given by

\be
\frac{dw}{dt}=\eta \mu -\eta \mu w^2
\label{eq:}
\ee
The intuitive behavior of this equation is clear. Suppose we start at, or near, $w=0$. Then the sign of the derivative at the origin is determined by the sign of $\mu$, and $w$ will either increase and  asymptotically converge towards +1 when $\mu >0$, or decrease and asymptotically converge towards -1 when $\mu <0$. Note also that
$w_{obv1}(t)=1$ and $w_{obv2}(t)=-1$ are two obvious constant solutions of the differential equation.

To solve the Riccati equation more formally we use the known obvious solutions and introduce the new variable $z=1/(w-w_{obv})=1/(w+1)$ (and similarly one can introduce the new variable $z=1/(w-1)$ to get a different solution). As a result,
$w=(1-z)/z$. It is then easy to see that the new variable $z$ satisfies a linear differential equation. More precisely, a simple calculation gives

\be
\frac{dz}{dt}=-2\eta \mu z + \eta \mu
\label{eq:}
\ee
resulting in

\be
z(t)=Ce^{-2 \eta \mu t}+{1 \over 2}
\label{eq:}
\ee
and thus

\be
w(t)=\frac{1-2Ce^{-2\eta \mu t}}{1+2Ce^{-2 \eta \mu t}} \quad {\rm with} \quad w(0)=\frac{1-2C}{1+2C} \quad {\rm or} \quad
C=\frac{1-w(0)}{2(1+w(0))}
\label{eq:}
\ee
Simulations are shown in Figure \ref{fig:riccatiunsup} for the unsupervised case, and Figure \ref{fig:riccatisup} for the supervised case.

\subsubsection{Oja's Rule}
An important example of a rule with $r>1$ is provided by Oja's rule \cite{oja1982simplified}

\be
\Delta w_i =\eta(OI_i-O^2w_i)
\label{eq:}
\ee
with $d=3$, originally derived for a linear neuron. The idea behind this rule is to control the growth of the weights induced by the simple Hebb rule 
by adding a decay term. The form of the decay term can easily be obtained by requiring the weights to have constant norm and expanding the corresponding constraint in Taylor series with respect to $\eta$. It can be shown under reasonable assumptions that the weight vector will converge towards the principal eigenvector of the input correlation matrix. 
Converging learning rules are discussed more broadly in Section \ref{sec:NewRules}.

\section{Local Learning in the Non-Linear Case}
\label{sec:NonLinear}
To extend the theory to the non-linear case, we consider a non-linear unit
$O=f(S)=f(\sum_0^N w_iI_i)$ where $f$ is a transfer function that is logistic (0,1) or hyperbolic tangent (-1,1) in the differentiable case, or the corresponding threshold functions. All the expectations computed in the linear case that do not involve the variable $O$ can be computed exactly as in the linear case. Furthermore, at least in the case of threshold gates, we can easily deal with powers of $O$ because $O^2=O$ in the (0,1) case, and $O^2=1$ in the (-1,1) case.
Thus, in essence, the main challenge is to compute terms of the form $E(O)$ and $E(OI_i)$ when $O$ is non-linear. We next show how these expectations can be approximated. 

\subsection{Terms in $E(O)$ and the Dropout Approximation}

When the transfer function is a sigmoidal logistic function $\sigma=\sigma_{[0,1]}$, we can use the approximation \cite{baldidropout14}

\be
E(O)=E(\sigma(S)) \approx \sigma(E(S)) \quad {\rm with} \quad 
\vert E(\sigma(S)) - \sigma(E(S)) \vert \approx
\frac{V\vert 1-2E \vert}{1-2V} \leq 2E(1-E)\vert 1-2E\vert
\label{eq:dropout1}
\ee
where $E=E(O)=E(\sigma(S))$ and $V={\rm Var}(O)={\rm Var}(\sigma(S))$. Thus 
$E(O)\approx \sigma(\sum_iw_i\mu_i)$. During learning, as the $w_i$ vary, this term could fluctuate. Note however that if the data is centered ($\mu_i=0$ for every $i$), which is often done in practice, then we can approximate the term $E(O)$ by a constant equal to $\sigma(0)$ across all epochs.
Although there are cases where the approximation of Equation \ref{eq:dropout1}
is not precise, in most reasonable cases it is quite good.
This approximation has its origin in the dropout approximation 
$E(O) \approx NWGM(O)=\sigma (E(S))$ where $NWGM$ represents the normalized geometric mean. These and several other related results are proven in  \cite{baldidropout14}.

When the transfer function is a hyperbolic tangent function we can use the same approximation 

\be
E(O)=E(\tanh (S)) \approx \tanh (E(S))
\label{eq:}
\ee
This is simply because

\be 
\tanh(S)= 2 \sigma(2S)-1
\label{eq:}
\ee
Equation \ref{eq:dropout1} is valid not only for the standard logistic function, but also for any logistic function with slope $\lambda$ of the form $\sigma(S)=1/(1+c e^{-\lambda S})$. Threshold functions are approximated by sigmoidal functions with $\lambda \to +\infty$. Thus the approximation can be used also for threshold functions with
$\lambda \to +\infty$, with similar caveats. More generally, if the transfer function $f$ is differentiable and can be expanded as a Taylor series around the mean $E(S)$, we always have:
$f(S)\approx f(E(S)) + f'(E(S)) (S-E(S)) +{1 \over 2} f''(E(S)) (S-E(S))^2$ and thus 
$E(f(S))\approx f(E(S)) + {1 \over 2} f''(E(S)) Var S$. Thus if $Var S$ is small or $f''(E(S))$ is small, then $E(f(S)) \approx f(E(S))=f(\sum_i w_i\mu_i)$. The approximations can often be used also for other functions (e.g. rectified linear), as discussed in \cite{baldidropout14}. 

\subsection{Terms in $E(OI_i)$ }

Next, in the analysis of learning rules in the non-linear case, we must deal with expectations of the form $E(OI_i)$. A first simple approximation is to assume that $O$ and $I_i$ are almost independent and therefore

\be
E(OI_i)\approx E(O)E(I_i)=E(O)\mu_i
\label{eq:}
\ee
In this expression, $E(O)$ can in turn be approximated using the method above. For instance, 
in the case of a logistic or $\tanh$ transfer function

\be
E(OI_i)\approx E(O)E(I_i)=E(O)\mu_i\approx \mu_i \sigma(E(S))=\mu_i \sigma(\sum_{i=1}^N w_i \mu_i)
\label{eq:}
\ee
If the data is centered, the approximation reduces to 0.

A second possible approximation is obtained by expanding the sigmoidal transfer function into a Taylor series.
To a first order, this gives

\be
E(OI_i)=E \left [ \sigma (\sum_j w_jI_j)I_i \right ]=E \left [ \sigma (\sum_{j \not = i} w_j I_j + w_iI_i)I_i \right ]\approx E \left [\sigma (\sum_{j \not = i} w_j I_j)I_i + \sigma '(\sum_{j \not = i} w_j I_j)w_iI_iI_i
\right ]
\label{eq:nl1}
\ee
with the approximation quality of a first-order Taylor approximation to $\sigma $.
To further estimate this term we need to assume that 
the terms depending on $j$ but not on $i$ are independent of the terms dependent on $i$, i.e. that the data covariances are 0. In this case,

\be
E(OI_i) \approx  E(\sigma (\sum_{j \not = i} w_j I_j))E(I_i) + E(\sigma '(\sum_{j \not = i} w_j I_j)w_iE(I_i^2)
)\approx \mu_i \sigma (\sum_{j \not = i}w_j \mu_j) + E(\sigma '(\sum_{j \not = i} w_j I_j))w_i(\mu_i^2 + \sigma_i^2)
\label{eq:}
\ee
where $\sigma_i^2=Var I_i$ and the latter approximation uses again the dropout approximation. If in addition the data is centered ($\mu_i=0$ for every $i$) we have 

\be
E(OI_i) \approx E(\sigma '(\sum_{j \not = i} w_j I_j))w_i \sigma_i^2
\label{eq:}
\ee
which reduces back to a linear term in $w$

\be
E(OI_i) \approx E(\sigma '(0))w_i \sigma_i^2
\label{eq:}
\ee
when the weights are small, the typical case at the beginning of learning.

\par
\null\par
In summary, when $n \leq 2$ and $r \leq 1$ the dynamics of the learning rule can be solved exactly or approximately, even in the non-linear case, and  it is entirely determined by the statistical moments of the data. 

\subsection{Examples}

In this section we consider simple local learning rules applied to a single sigmoidal or threshold unit.

\subsubsection{Unsupervised Simple Hebb Rule}

\begin{figure}[h!]
    \centering
    \includegraphics[width=0.8\textwidth]{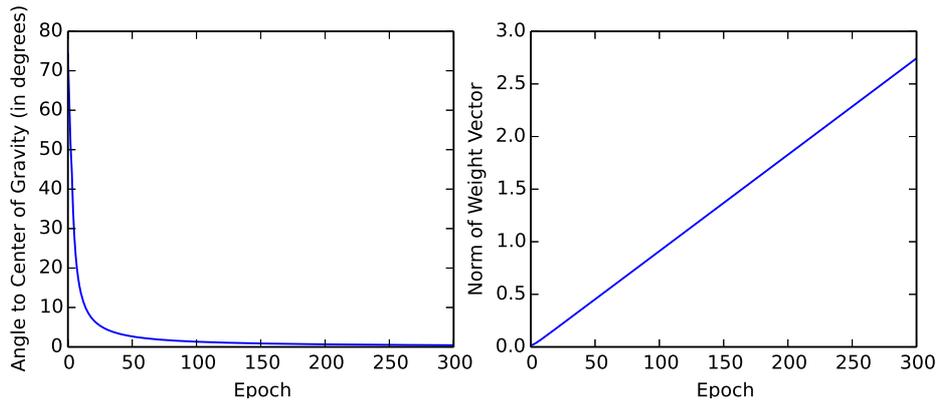}
    \caption{Single unit trained on the MNIST data set (60,000 examples) for 500 epochs, with a learning rate of 0.001 using the simple Hebb rule in unsupervised fashion. The fan-in is 784 ($28 \times 28$). The weights are initialized from a normal distribution with standard deviation 0.01.
    Left: Angle of the weight vector to the center of gravity. Right: Norm of the weight vector.}
    \label{fig:hebbanglenorm}
\end{figure}

We first consider the simple Hebb rule $\Delta w_i =\eta I_iO$. Using the approximations described above we obtain

\be
E(\Delta w_i) \approx \eta \mu_i E(O) \approx \eta  \mu_i \sigma (\sum_i w_i \mu_i)
\quad {\rm thus} \quad
w_i(k)=w_i(0)+\eta\mu_i \left [\sum_{l=0}^{k-1} \sigma(\sum_j w_j(l)\mu_j) \right ]
\label{eq:}
\ee
Thus the weight vector tends to align itself with the center of gravity of the data.
However, this provides only a direction for the weight vector which continues to grow to infinity along that direction, as demonstrated in Figure \ref{fig:hebbanglenorm}.

\subsubsection{Supervised Simple Hebb Rule}

Here we consider a single [-1,1] sigmoidal or threshold unit trained using a set of $M$ training input-output pairs $I(t),T(t)$ where $T(t)=\pm 1$.
In supervised mode, here the output is clamped to the target value (note in general this is different from the perceptron or backpropagation rule). 

Here we apply the simple Hebb rule with the output clamped to the target so that 
$\Delta w_i=\eta I_iT$.
Thus the expectation $E(\Delta w_{i})=\eta E(I_iT)$ is constant across all the epochs and depends only on the data moments. In general the weights will grow linearly with the number of epochs, unless $E(\Delta w_{i})=0$
in which case the $w_i$ will remain constant and equal to the initial value $w_i(0)$. In short,

\be
w_i(k)= w_i(0) + k \eta E(I_iT)
\label{eq:}
\ee
If the targets are essentially independent of the inputs,
we have
$E(\Delta w_i) \approx \eta E(I_i)E(T)=\eta \mu_i E(T)$,
and thus after $k$ learning epochs the weights are given by 

\be
w_i(k)= w_i(0) + k \eta  \mu_i \mu_T
\label{eq:}
\ee
In this case we see again that the weight vector tends to be co-linear with the center of gravity of the data, with a sign that depends on the average target, and a norm that grows linearly with the number of epochs.

\subsubsection{Gradient Descent Rule}

A last example of a convergent rule is provided by 
$\Delta w_i = \eta (T-O)I_i$ with the logistic transfer function. The rule is convergent (with properly decreasing learning rate $\eta$) because it performs gradient descent on the relative entropy error function 
$E_{err}(w)=- \sum_{t=1}^M T(t)\log O(t)+(1-T(t))\log(1-O(t))$. 
Remarkably, up to a trivial scaling factor of two that can be absorbed into the learning rate, this learning rule has exactly the same form when the $\tanh$ function is used over the $[-1,1]$ range (Appendix B). 

\section{Derivation of New Learning Rules}
\label{sec:NewRules}
The local learning framework is also helpful for discovering new learning rules. In principle, one could recursively enumerate all polynomial learning rules with rational coefficients and search for rules 
satisfying particular properties. However this is not necessary for several reasons. In practice, we are only interested in polynomial learning rules with relatively small degree (e.g. $n \leq 5$) and more direct approaches are possible. To provide an example, here we consider the issue of convergence and derive new convergent learning rules.

We first note that a major concern with a Hebbian rule, even in the simple case 
$\Delta w_{ij} \propto O_iO_j$, is that the weights tend to diverge over time towards very large positive or very large negative values. To ensure that the weights remain within a finite range, it is natural to introduce a decay term so that  $\Delta w_{ij} \propto O_iO_j - Cw_{ij}$ with $C>0$. The decay coefficient can also be adaptive as long as it remains positive. This is exactly what happens in Oja's cubic learning rule
 \cite{oja1982simplified} 

\be 
\Delta w_{ij} \propto O_iO_j - O_i^2 w_{ij}
\label{eq:oja10}
\ee
which has a weight decay term $O_i^2 w_{ij}$ proportional to the square of the output and is known to extract the principal component of the data. Using different adaptive terms, we immediately get new rules such as:

\be 
\Delta w_{ij} \propto O_iO_j - O_j^2 w_{ij}
\label{eq:oja11}
\ee
and

\be 
\Delta w_{ij} \propto O_iO_j - (O_iO_j)^2 w_{ij}=O_iO_j(1-O_iO_jw_{ij})
\label{eq:oja12}
\ee
And when the postsynaptic neuron has a target $T_i$, we can consider the clamped or gradient descent version of these rules. In the clamped cases, some or all the occurrences of $O_i$ in Equations \ref{eq:oja10},
\ref{eq:oja11}, and \ref{eq:oja12} are to be replaced by the target $T_i$.
In the gradient descent version, some or all the
occurrences of $O_i$ in Equations \ref{eq:oja10},
\ref{eq:oja11}, and \ref{eq:oja12} are to be replaced by  $(T_i-O_i)$.
The corresponding list of rules is given in Appendix C.

To derive additional convergent learning rules, we can take yet a different approach by
introducing a saturation effect on the weights. To ensure that the weights remain in the $[-1,1]$ range, we can assume that the weights are calculated by applying a hyperbolic tangent function. 

Thus consider a [-1,1] system trained using the simple Hebb rule $\Delta w_{ij} \propto O_iO_j$. To keep the weights in the [-1,1] range throughout learning, we can write:  

\be
w_{ij}(t+1)= \tanh[w_{ij}(0)+ \eta O_i(1)O_j(1)+ \ldots  \eta O_i(t)O_j(t)+\eta O_i(t+1)O_j(t+1)]
\label{eq:newrule1}
\ee
where $\eta$ is the learning rate. By taking a first order Taylor expansion and using the fact that $\tanh(x)'=1-\tanh^2(x)$, we obtain the new rule 

\be
w_{ij}(t+1)=w(t)+\eta(1-w_{ij}^2) O_i(t) O_j(t) \quad {\rm or} \quad
\Delta w_{ij} \propto (1-w_{ij}^2)O_iO_j
\label{eq:NewRule10}
\ee
Note that while simple, this is a quartic learning rule
in the local variables with $n=4$ and $d=3$.
The rule forces $\Delta w_{ij} \to 0$ as
$\vert w_{ij} \vert \to 1$. 
In the supervised case, this rule becomes

\be
\Delta w_{ij} \propto (1-w_{ij}^2)T_iO_j
\label{eq:}
\ee
in the clamped setting, and

\be
\Delta w_{ij} \propto (1-w_{ij}^2)(T_i-O_i)O_j
\label{eq:}
\ee
in the gradient descent setting. 

\begin{figure}[h!]
    \centering
    \includegraphics[width=0.8\textwidth]{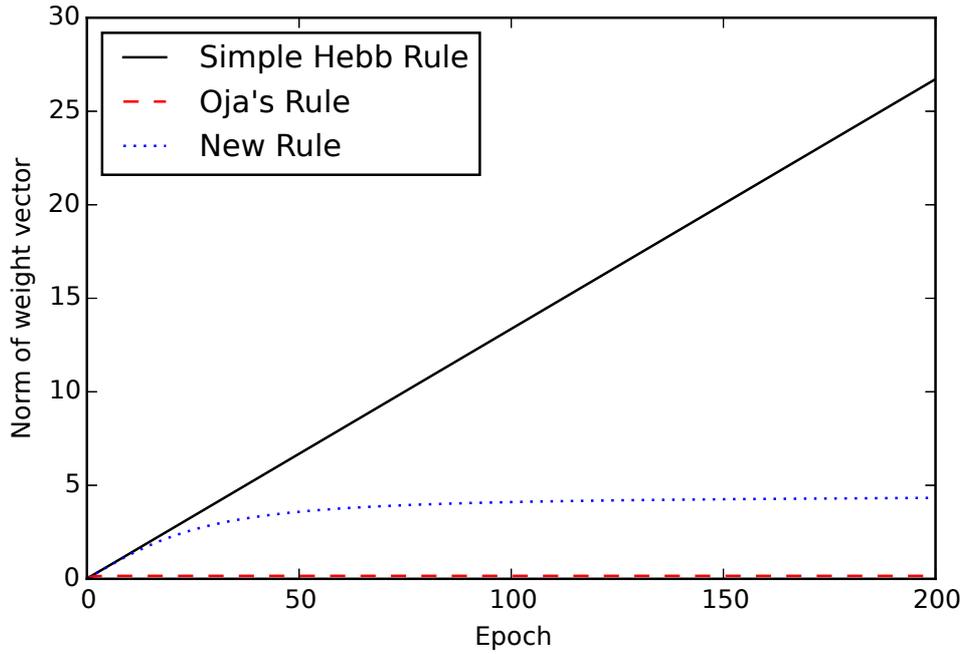}
    \caption{Temporal evolution of the norm of the weight vector of a single threshold gate with 20 inputs and a bias trained in supervised mode using 500 randomly generated training examples using three different learning rules:
Basic Hebb, Oja, and the New Rule. Oja and the New Rule gracefully prevent the unbounded growth of the weights.
The New Rule produces a weight vector whose component are fairly saturated (close to -1 or 1) with a total norm close to $\sqrt 21$. }
    \label{fig:weights}
\end{figure}

\begin{figure}[h!]
    \centering
    \includegraphics[width=0.8\textwidth]{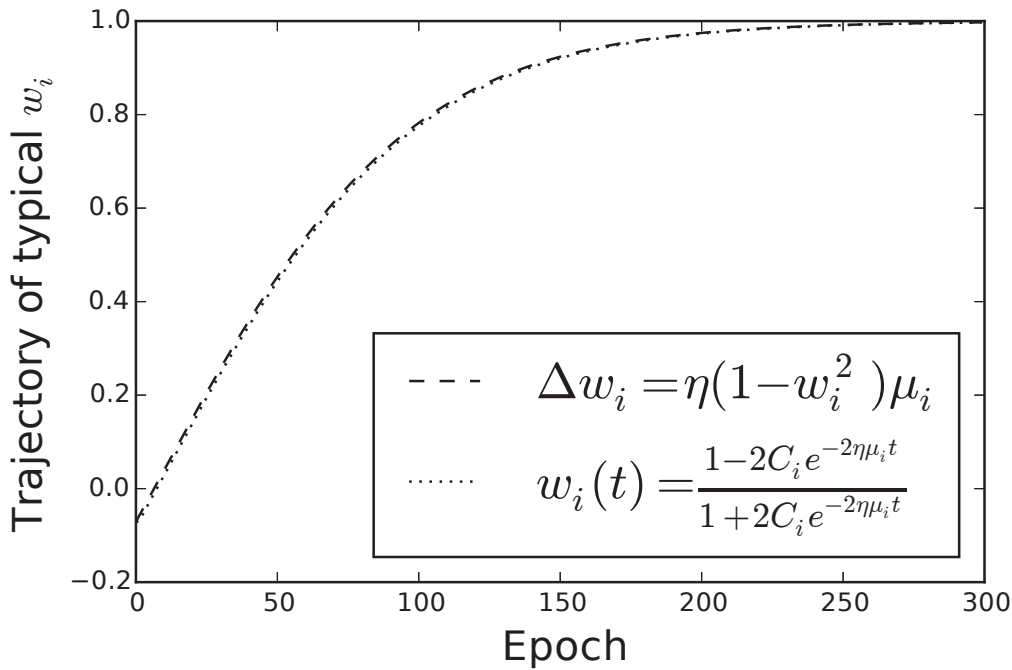}
    \caption{A learning rule that results in a Riccati differential equation. The solution to this Riccati equation tells us that all the weights will converge to 1. A typical weight is is shown.  It is initialized randomly from $N(0,0.1)$ and trained on 1000 MNIST  resulting in a fan-in of 784 ($28 \times 28$). There is almost perfect agreement between the theoretical and empirical curve.}
    \label{fig:riccatiunsup}
\end{figure}

\begin{figure}[h!]
    \centering
    \includegraphics[width=0.8\textwidth]{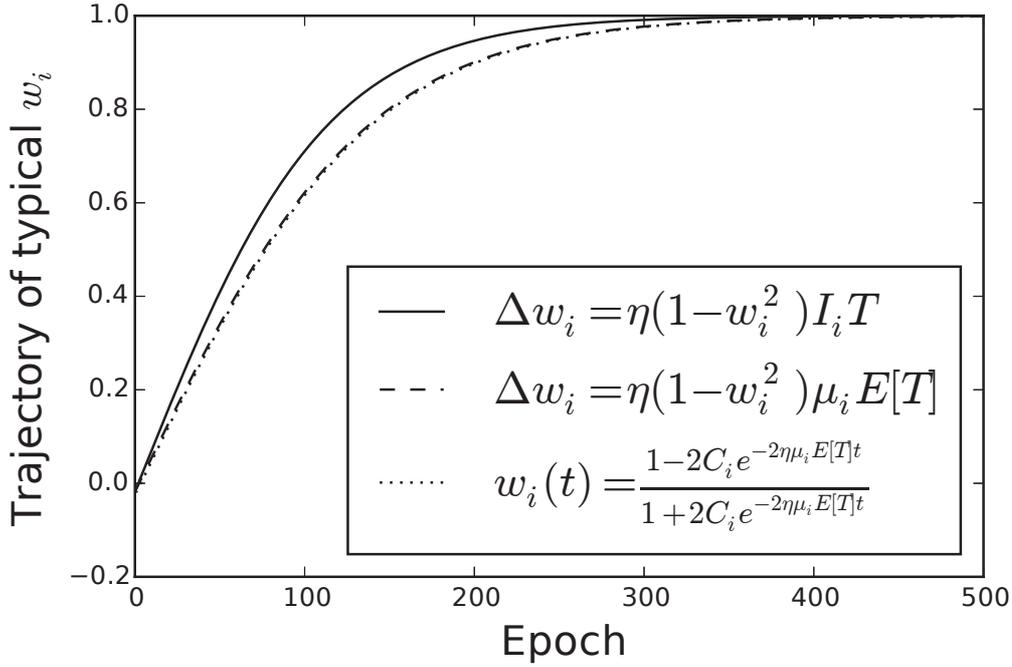}
    \caption{
        When the independence assumption is reasonable, the Riccati equation describes the dynamics of learning and can be used to find the exact solution. The typical weight shown here is randomly initialized from $N(0,0.1)$ and is trained on $M= 1000$ MNIST samples to recognize digits 0-8 vs 9 classes. $N=784$. }
    \label{fig:riccatisup}
\end{figure}

\begin{figure}[h!]
    \centering
    \includegraphics[width=0.8\textwidth]{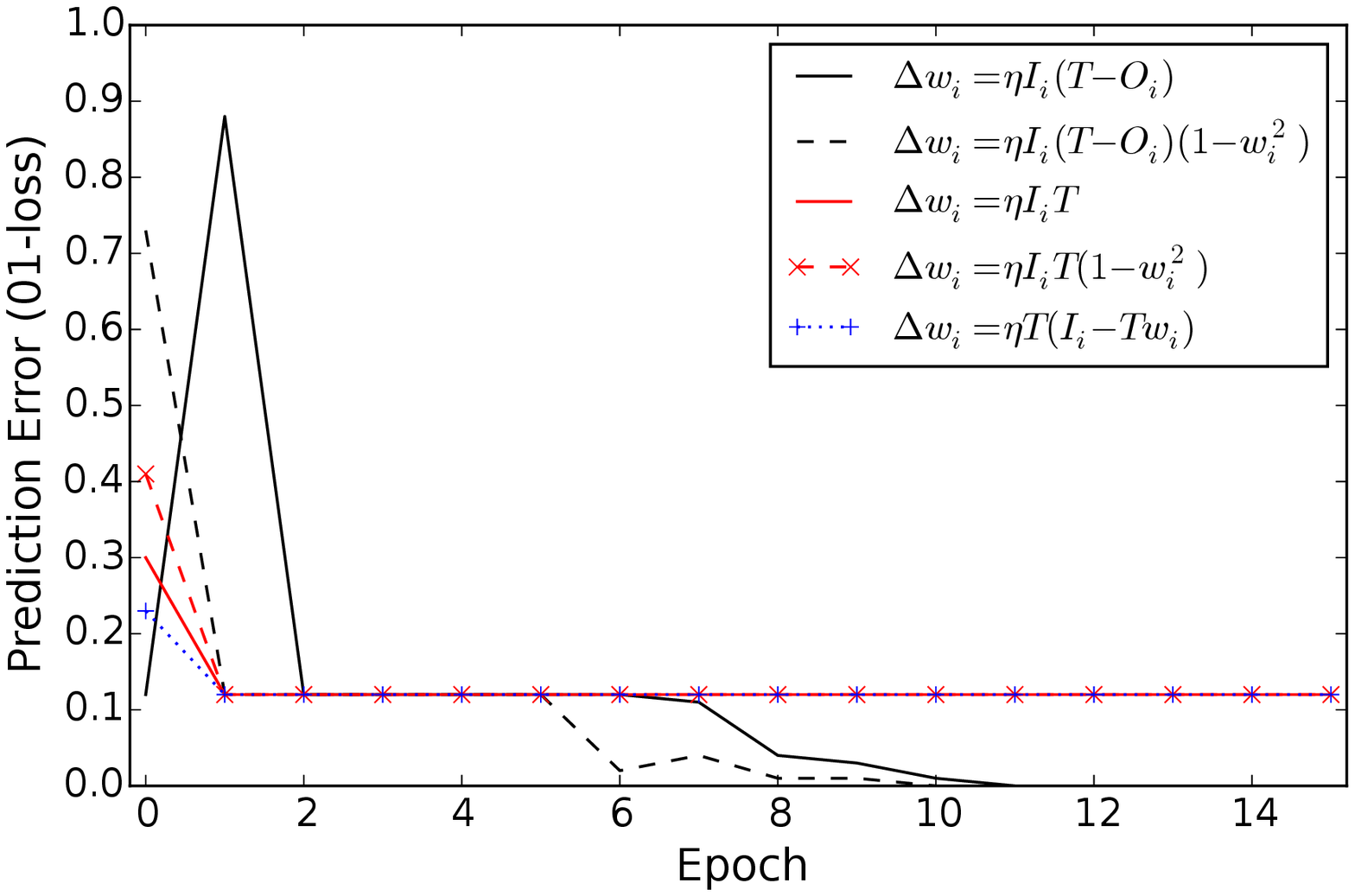}
    \caption{A single neuron with tanh activation trained to recognize the handwritten digit nine with five supervised learning rules. The input data is 100 MNIST images (made binary by setting pixels to +1 if the greyscale value surpassed a threshold of 0.2, and -1 otherwise), and binary {-1,+1} targets. Weights were initialized independently from N(0, 0.1), and updated with learning rate $\eta=0.1$. }
    \label{fig:supervised}
\end{figure}

To further analyze the behavior of this rule in the clamped setting, for instance, let us consider a single tanh or threshold  [-1,1] unit with
$\Delta w_i=\eta (1-w^2)TI_i$. In the regime where the independence approximation is acceptable, this yields
$E(\Delta w_i)=\eta (1-w^2)E(T) \mu_i$ which is associated with the Riccati differential equation that we already solved in the linear case. One of the solutions (converging to +1) is given by

\be
w(k)=\frac{1-2C e^{-2\eta \mu E(t) k}}{1+2C e^{-2\eta \mu E(t) k} }
\quad {\rm with} \quad  C=\frac{1-w(0)}{2(1+w(0))}
\label{eq:}
\ee
Simulations of these new rules demonstrating how they effectively control the magnitude of the weights and how well the theory fits the empirical data are shown in Figures \ref{fig:weights}, \ref{fig:riccatiunsup}, \ref{fig:riccatisup},
and \ref{fig:supervised}. 

Finally, another alternative mechanism for preventing unlimited growth of the weights is to reduce the learning rate as learning progresses, for instance using a linear decay schedule.

\section{What is Learnable by Shallow or Deep Local Learning}
\label{sec:Limitations}
The previous sections have focused on the study of local learning rules, stratified by their degree, in shallow networks. In this section, we begin to look at local learning rules applied to deep feedforward networks and partially address the question of what is locally learnable in feedforward networks.

Specifically we want to consider shallow (single adaptive layer) local learning, or {\it deep local learning} defined as learning in deep feedforward layered architectures with learning rules of the form $\Delta w_{ij} =F(O_i,O_j, w_{ij})$ applied successively to all the layers, starting from the input layer, possibly followed by supervised learning rule of the form $\Delta w_{ij} =F(O_i,O_j, T_i, w_{ij})$ in the top layer alone, when targets are available for the top layer (Figure \ref{fig:DeepLocal}). One would like to understand what input-output functions can be learnt from examples using this strategy and whether this provides a viable alternative to back-propagation.
We begin with a few simulation experiments to further motivate the analyses.

\subsection{Simulation Experiments: Learning Boolean Functions}

We conduct experiments using various local learning rules to try to learn Boolean functions with {\it small} fan in architectures with one or two adaptive layers. These experiments are purposely carried to show how simulations run in simple cases can raise false hopes of learnability by local rules that do not extend to large fan in and more complex functions, as shown in a later section.
Specifically, we train 
binary [-1,1] threshold gates to learn Boolean functions of up to 4 inputs,
using the simple Hebb rule, the Oja rule, and the new rule corresponding to Equation
\ref{eq:NewRule10} and its supervised version. Sometimes multiple random initializations of the weights are tried and the function is considered to be learnable if it is learnt in at least one case. [Note: The number of trials needed is not important since the ultimate goal is to show that this local learning strategy cannot work for more complex functions, even when a very large number of trials is used.]
In Tables \ref{tab:TableBoolean1} and \ref{tab:TableBoolean2}, 
we report the results obtained both in the shallow case (single adaptive layer) trained in a supervised manner, and the results obtained in the deep case (two adaptive layers) where the adaptive input layer is trained in unsupervised manner and the adaptive  output layer is trained in supervised manner. In the experiments, all inputs and targets are binary (-1,1), all the units have a bias, and the learning rate decays linearly.

As shown in Table \ref{tab:TableBoolean1}, 14 of the 16 possible Boolean functions of two variables ($N=2$) can be learnt using the Simple Hebb, Oja, and new rules. The two Boolean functions that cannot be learnt are of course XOR and its converse which cannot be implemented by a single layer network.
Using deep local learning in two-layer networks, then all three rules are able to learn all the Boolean functions with $N=2$, demonstrating that at least some complex functions can be learnt by combining unsupervised learning in the lower layer with supervised learning in the top layer. 
Similar results are also seen for 
$N=3$, where 104 Boolean functions, out of a total of 256, are learnable in a shallow network. And all 256 functions are learnable by a two-layer network by any of the three learning rules.

Table \ref{tab:TableBoolean2} shows similar results on the subset  
of {\it monotone} Boolean functions. As a reminder, a Boolean function is said to be monotone if increasing the total number of +1 in the input vector can only leave the value of the output unchanged or increase its value from -1 to +1. Equivalently, it is the set of Boolean functions with a circuit comprising only AND and OR gates. There are recursive methods for generating monotone Boolean functions and the total number of monotone Boolean functions is known as the Dedekind number. For instance, there are 168 monotone Boolean functions with $N=4$ inputs. Of these, 150 are learnable by a single unit trained in supervised fashion, and all 168 are learnable by a two-layer network trained with a combination of unsupervised (input layer) and supervised (output layer) application of the three local rules.

\begin{table}[h!]
\renewcommand{\arraystretch}{1.5}
  \begin{center}
  \begin{tabular}{ | c | c | c | c | c | c }
    \hline
    \textbf{Fan } & \multicolumn{2}{c|}{\textbf{Functions Learnt}} &  \textbf{Total Number } & \textbf{Rule} \\ 
    \textbf{In } & \textbf{Shallow} & \textbf{Deep} &  \textbf{of Functions} & \textbf{}
        \\ \hline  
    2 & 14 &16 & 16 & Simple Hebb \\ \hline
    2 & 14 &16  & 16 & Oja \\ \hline
    2 & 14 &16  & 16 & New \\ \hline
    3 & 104 &256& 256  & Simple Hebb \\ \hline
    3 & 104 &256 & 256  & Oja \\ \hline
   	3 & 104 &256& 256  & New \\ \hline
    \end{tabular}  
       \end{center} 
\caption{Small fan-on Boolean functions learnt by deep local learning.}
    \label{tab:TableBoolean1}
\end{table}

\begin{table}[h!]
\renewcommand{\arraystretch}{1.5}
  \begin{center}
    \begin{tabular}{ | c | c | c | c | c | c }
    \hline
    \textbf{Fan } & \multicolumn{2}{c|}{\textbf{Functions Learnt}} &  \textbf{Total Number } & \textbf{Rule} \\ 
    \textbf{In } & \textbf{Shallow} & \textbf{Deep} &  \textbf{of Functions} & \textbf{}
        \\ \hline
    2 & 6 &6 & 6 & Simple Hebb \\ \hline
    2 & 6& 6 & 6 & Oja \\ \hline
    2 & 6 &6 & 6 & New \\ \hline
    3 & 20 &20 & 20 & Simple Hebb \\ \hline
    3 & 20 &20 & 20 & Oja \\ \hline
   	3 & 20 &20 & 20 & New \\ \hline
    4 & 150& 168 & 168 & Simple Hebb\\ \hline
    4 & 150 &168 & 168 & Oja\\ \hline
    4 & 150 &168 & 168 & New\\ \hline
    \end{tabular}
       \end{center} 
\caption{Small fan-in monotone Boolean functions learnt by deep local learning.}
    \label{tab:TableBoolean2}
\end{table}

In combination, these simulations raise the question of what are the classes of functions learnable by shallow or deep local learning,  and raise the (false) hope that purely local learning may be able to replace backpropagation.

\subsection{Learnability in Shallow Networks}

Here we consider in more detail the learning problem for a single [-1,1] threshold gate, or perceptron.

\subsubsection{Perceptron Rule}
 In this setting, the problem has already been solved at least in one setting by the perceptron learning algorithm and theorem 
\cite{rosenblatt1958perceptron,minsky1969perceptron}. 
Obviously, by definition, a threshold gate can only implement in an exact way functions (Boolean or continuous) that are {\it linearly separable}. The perceptron learning algorithm simply states that if the data is linearly separable, the local gradient descent learning rule
$\Delta w_i = \eta (T-O)I_i$ will converge to such a separating hyperplane. Note that this is true also in the case of [0,1] gates
as the gradient descent rule as the same form in both systems.
When the training data is not linearly separable, the perceptron algorithm is still well behaved in the sense that algorithm converges to a relatively small compact region \cite{minsky1969perceptron,block1970boundedness,gelfand2010herding}.
Here we consider similar results for a slightly different supervised rule, the clamped form of the simple Hebb rule: $\Delta w_i = \eta TI_i$ .

\subsubsection{Supervised Simple Hebb Rule}

Here we consider a supervised training set consisting of input-target pairs of the form
${\cal S}=\{(I(t),T(t)): t=1, \ldots, M \}$ where the input vectors $I(t)$ are $N$-dimensional (not-necessarily binary) vectors with corresponding targets $T(t)=\pm1$ for every $t$
(Figure \ref{fig:ssh}). 
$\cal S$ is linearly separable (with or without bias) if there is a separating hyperplane, i.e. set of weights $w$ such that $\tau(I(t))=\tau (\sum w_i I_i(t))=T(t)$ (with or without bias) for every $t$, where $\tau$ is the $\pm 1$ threshold function. 
To slightly simplify the notation and analysis, throughout this section, we do not allow ambiguous cases where $\tau(I)=0$ for any $I$ of interest. In this framework, the linearly separable set $\cal S$ is learnable by a given learning rule R (R-learnable) if the rule can find a separating hyperplane. 

\par\null
\noindent
{\bf The Case Without Bias:}
When there is no bias ($w_0=0$), then $\tau(-I)=-\tau (I)$ for every $I$.
In this case, a set $\cal S$ is {\it consistent} if for every $t_1$ and $t_2$: $I(t_1)=-I(t_2) \implies T(t_1)=-T(t_2)$. Obviously consistency is a necessary condition for separability and learnability in the case of 0 bias. 
When the bias is 0, the training set $S$ can be put into its {{\it canonical} form ${\cal S}^c$ by ensuring that all targets are set to $+1$, replacing any training pair of the form $(I(t), -1)$ by the equivalent pair $(T(t)I(t),+1)=(-I(t), +1)$. Thus the size of a learnable canonical training set in the binary case, where $I_i(t)=\pm 1$ for every $i$ and $t$, is at most $2^{N-1}$. 

We now consider whether $\cal S$ is learnable by the supervised simple Hebb rule (SSH-learnable) corresponding to clamped outputs 
$\Delta w_i=\eta I_iT$, first in the case where there is no bias, i.e. $w_0=0$.  We let $Cos$ denote the $M\times M$ symmetric square matrix of cosine values $Cos=(Cos_{uv})=(\cos (T(u)I(u),T(v)I(v)))=
(\cos (I^c(u),I^c(v)))$.
It is easy to see that applying the supervised simple Hebb rule with the vectors in $\cal S$ is equivalent to applying the supervised simple Hebb rule with the vectors in ${\cal S}^c$, both leading to the same weights.
If $\cal S$ is in canonical form and there is no bias, we have the following properties.
\par
\null\par
\noindent
{\bf Theorem:}
\begin{enumerate}
\item The supervised simple Hebb rule
leads to $\Delta w_i=\eta E(I_iT)=\eta E(I^c_i)=\eta \mu^c_i$ and thus
$w(k)=w(0)+\eta k \mu^c$.
\item A necessary condition for $\cal S$ to be SSH-learnable is that $\cal S$ (and equivalently ${\cal S}^c$ ) be linearly separable by a hyperplane going through the origin.
\item A sufficient condition for  $\cal S$ to be SSH-learnable from any set of starting weights is that all the vectors in ${\cal S}^c$ be in a common orthant, i.e. that the angle between any $I^c(u)$ and $I^c(c)$ lie between $0$ and $\pi/2$ or, equivalently, that 
$ 0 \leq \cos(I^c(u),I^c(v)) \leq 1$ for any $u$ and $v$.
\item A sufficient condition for  $\cal S$ to be SSH-learnable from any set of starting weights is that all the vectors in $\cal S$(or equivalently in ${\cal S}^c$) be orthogonal to each other, i.e. $I(u)  I(v)=0$ for any $u \neq v$.
\item If all the vectors $I(t)$ have the same length, in particular in the binary $\pm 1$ case,
$S$ is SSH-learnable from any set of initial weights if and only if the sum of any row or column of the cosine matrix associated with ${\cal S}^c$ is strictly positive.
\end{enumerate}

\noindent
{\bf Proof:}
\par\noindent
1) Since ${\cal S}^c$ is in canonical form, all the targets are equal to +1, and thus $E(I_i(t)T(t))=E(I^c_i(t))=\mu^c_i$. After $k$ learning epochs, 
with a constant learning rate, the weight vector is given by $w(k)=w(0)+\eta k \mu^c$. 
\par\noindent
2) This is obvious since the unit is a threshold gate.
\par \noindent
3) For any $u$, the vector $I^(u)$ has been learnt after $k$ epochs if and only if
\be
\sum_{i=1}^N [w_i(0) +\eta k E(I_i(t)T(t))]I_i(u)=
\sum_{i=1}^N \left ( w_i(0)I^c_i(u) +\eta k \frac{1}{M} \sum_{t=1}^M I^c_i(t)I^c_i(u)\right ) >0
\label{eq:proof3}
\ee
Here we assume a constant positive learning rate, so after a sufficient number of epochs the effect of the initial conditions on this inequality can be ignored.  Alternatively one can examine the regime of decreasing learning rates using initial conditions close to 0.
Thus ignoring the transient effect caused by the initial conditions, 
and separating the terms corresponding to $u$,
$I(u)$ will be learnt after a sufficient number of epochs if and only if

\begin{multline}
\sum_{i=1}^N \sum_{t=1}^MI^c_i(t)I^c_i(u)=\sum_{t=1}^M I^c(t)I^c(u)=
\sum_{t=1}^M \vert \vert I^c(t)\vert \vert \vert \vert I^c(u) \vert \vert
\cos(I^c(t),I^c(u))\\
=\vert I^c(u) \vert^2+ \sum_{t \neq u} \vert \vert I^c(t)\vert \vert \vert \vert I^c(u) \vert \vert
\cos(I^c(t),I^c(u))>0
\label{eq:proof4}
\end{multline}
Thus if all the cosines are between 0 and 1 this sum is strictly positive (note that we do not allow $I(u)=0$ in the training set).
Since the training set is finite, we simply take the maximum number of epochs over all training examples where this inequality is satisfied, to offset the initial conditions. Note that the expression in Equation
\ref{eq:proof4} is invariant with respect to any transformation that preserves vector lengths and angles, or changes the sign of all or some of the angles. Thus it is invariant with respect to any rotations, or symmetries.
\par \noindent
4) This is a special case of 3, also obvious from Equation \ref{eq:proof4}. Note in particular that a set of $\alpha N$ ($0<\alpha \leq 1$)  vectors chosen randomly (e.g. uniformly over the sphere or with fair coin flips) will be essentially orthogonal and thus learnable with high probability when $N$ is large.
\par \noindent
5) If all the training vectors have the same length $A$ (with $A=\sqrt N$ in the binary case), Equation \ref{eq:proof4} simply becomes

\be
A^2\sum_{t=1}^M \cos (I^c(u),I^c(t)) >0
\label{eq:}
\ee
and the property is then obvious. Note that it is easy to construct counterexamples where this property is not true if the training vectors do not have the same length. Take, for instance,
${\cal S}=\{(I(1),+1),(I(2),+1) \}$ with $I(1)=(1,0,0,\ldots,0)$ and
$I(2)=(-\epsilon,0,0 \dots,0)$ for some small $\epsilon>0$.
\par \noindent

\par\null
\noindent
{\bf The Case With Adaptive Bias:} 
When there is a bias ($w_0$ is not necessarily 0), starting from the training set ${\cal S}=\{(I(t),T(t)) \}$ we first modify
each vector $I(t)$ into a vector $ I'(t)$  by adding a zero-th component equal to +1,
so that $I'_0(t)=+1$, and $I'_i(t)=I_i(t)$ otherwise. Finally, we construct the corresponding canonical set ${\cal S}^c$ as in the case of 0 bias by letting 
${\cal S}^c=\{( I^c(t),+1) \}=\{(T(t)I'(t),+1) \}$ and apply the previous results to 
${\cal S}^c$. It is easy to check that applying the supervised simple Hebb rule with the vectors in $\cal S$ is equivalent to applying the supervised simple Hebb rule with the vectors in ${\cal S}^c$, both leading to the same weights.

\par
\null\par
\noindent
{\bf Theorem:}
\begin{enumerate}
\item The supervised simple Hebb rule applied to ${\cal S}^c$
leads to $\Delta w_i=\eta E(I'_iT)=\eta E(I^c_i)= \eta\mu^c_i$ and thus
$w(k)=w(0)+\eta k \mu^c$. The component $\mu_0^c$ is equal to the proportion of vectors in $\cal S$ with a target equal to +1.
\item A necessary condition for $\cal S$ to be SSH-learnable is that ${\cal S}^c$ be linearly separable by a hyperplane going through the origin in $N+1$ dimensional space.
\item A sufficient condition for  $\cal S$ to be SSH-learnable from any set of starting weights is that all the vectors in ${\cal S}^c$ be in a common orthant, i.e. that the angle between any $I^c(u)$ and $I^c(v)$ lie between $0$ and $\pi/2$ or, equivalently, that 
$ 0 \leq \cos(I^c(u),I^c(v)) \leq 1$ for any $u$ and $v$.
\item A sufficient condition for  $\cal S$ to be SSH-learnable from any set of starting weights is that all the vectors in ${\cal S}^c$ be orthogonal to each other, i.e. $I^c(u) I^c(v)=0$ for any $u \neq v$.
\item If all the vectors $I^c(t)$ have the same length, in particular in the binary $\pm 1$ case,
$S$ is SSH-learnable from any set of initial weights if and only if the sum of any row or column of the cosine matrix $Cos=(\cos(I^c(u),I^c(v)))$
is strictly positive.
\end{enumerate}

\noindent
{\bf Proof:}
\par\noindent
The proofs are the same as above. 
Note that $\mu_0^c=E(I_0^c(t))=E(I_0'(t)T(t))=E(T(t))$. 

\begin{figure}[h!]
    \centering
    \includegraphics[width=0.8\textwidth]{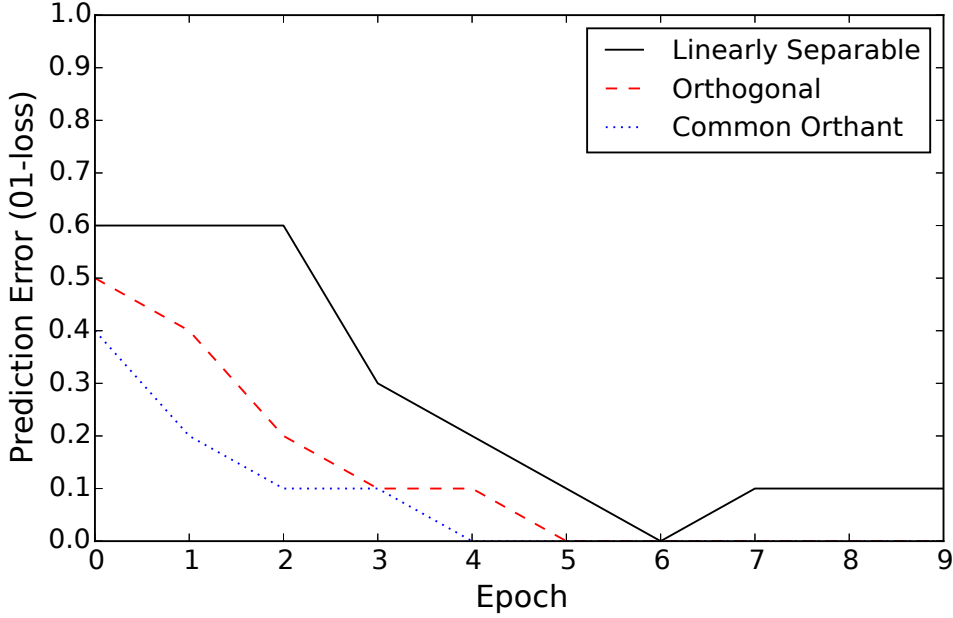}
    \caption{Examples of supervised simple Hebb learning with different training set properties. The linearly separable data is a random matrix of binary {-1,+1} values (p=0.5) of shape M=10, N=10, with binary {-1,1} targets determined by a random hyperplane. The orthogonal dataset is simply the identity matrix (multiplied by the scalar $\sqrt{10}$) and random binary {-1,+1} targets. The common orthant dataset was created by sampling the features in each column from either $[-1,0)$ or $(0,1]$, then setting all the targets to +1.  The weights were initialized independently from N(0,1), and weights were updated with the learning rate $\eta=0.1$. }
    \label{fig:ssh}
\end{figure}

\subsection{Limitations of Shallow Local Learning}

In summary, strictly local learning in a single threshold gate or sigmoidal function can learn any linearly separable function. While it is as powerful as the unit allows it to be, this form of learning is limited in the sense that it can learn only a very small fraction of
all possible functions. 
This is because the logarithm of the size of the set of all possible Boolean functions of $N$ variables is exponential and equal to $2^N$, whereas the logarithm of the size of the total number of linearly separable Boolean functions scales polynomially like $N^2$. Indeed, the total number $T_N$ of threshold functions of $N$ variables satisfies

\be
 N(N-1)/2\leq\log_2  T_N \leq {N^2}
 \label{eq:}
 \ee
 (see \cite{zuev1989asymptotics,cover1965geometrical,
muroga1965lower,baldi88a} and references therein). 
The same negative result holds also for the more restricted class of monotone Boolean functions, or any other class of exponential size. Most monotone Boolean functions cannot be learnt by a single linear threshold unit because
the number $M_N$ of monotone Boolean functions of $N$ variables, known as the Dedekind number, satisfies \cite{kleitman1969dedekind}

\be
 {N \choose \lfloor N/2 \rfloor }\leq \log_2 M_N \leq
{N \choose \lfloor N/2 \rfloor } \left (1 + O(\log N/N) \right )
\label{eq:}
\ee

These results are immediately true also for polynomial threshold functions, where the polynomials have bounded degree, by similar counting arguments \cite{baldi88a}. {\it In short, linear or bounded-polynomial threshold functions can at best learn a vanishingly small fraction of all Boolean functions, or any subclass of exponential size, regardless of the learning rule used for learning.}

The fact that local learning in shallow networks has significant limitations seems to be a consequence of the limitations of shallow networks, which are simply not able to implement complex function. This alone, does not preclude the possibility that iterated shallow learning applied to deep architectures, i.e. deep local learning, may be able to learn complex functions. After all this would be consistent with what is observed in the simple simulations described above where the XOR function, which is not learnable by a shallow networks, becomes learnable by local rules in a network of depth two. Thus over the years many attempts have been made to seek efficient, and perhaps more biologically plausible, alternatives to backpropagation for learning complex data using {\it only local rules}. For example, in one of the simplest cases, one could try to learn a simple two-layer autoencoder using unsupervised local learning in the first layer and supervised local learning in the top layer. More broadly, one could for example try to learn the MNIST benchmark \cite{lecun_gradient-based_1998} data using purely local learning. Simulations show (data not shown) however that such schemes fail regardless of which local learning rules are used, how the learning rates and other hyperparameters are tuned, and so forth. In the next section we show why all the attempts that have been made in this direction are bound to fail. 

\subsection{Limitations of Deep Local Learning}

Consider now deep local learning in a deep layered feedforward  architecture (Figure \ref{fig:DeepLocal}) with
$L+1$ layers of size
$N_0,N_1, \ldots N_L$ where layer 0 is the input layer, and layer $L$ is the output layer. We let $O_i^h$ denote the activity of unit $i$ in layer $h$ with $O_i^h=f(S_i^h)=f(\sum_j w_{ij}^h O_j^{h-1})$. The non-linear processing units can be fairly arbitrary. For this section, it will be sufficient to assume that the functions $f$ be differentiable functions of their synaptic weights and inputs. It is also possible to extend the analysis to, for instance, threshold gates by taking the limit of very steep differentiable sigmoidal functions.
We consider the supervised learning framework with a training set of input-output vector pairs of the form
$(I(t),T(t))$ for $t=1, \ldots, M$ and the goal is to minimize a differentiable error function $E_{err}$.
The main learning constraint is that we can only use deep local 
learning (Figure \ref{fig:DeepLocal}).

\begin{figure}[h!]
\begin{center}
\includegraphics[width=0.9\columnwidth]{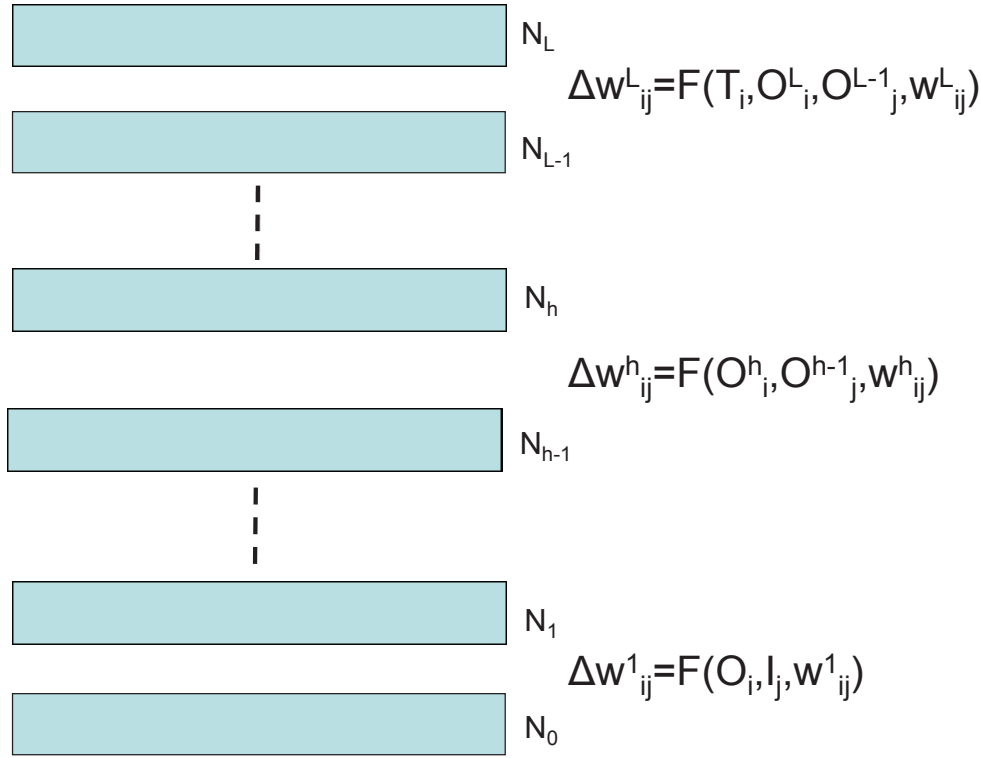}
\end{center}
\caption{\small Deep local learning. Local learning rules are used for each unit. For all the hidden units, the local learning rules are unsupervised and thus of the form
$\Delta w^h_{ij}=F(O^h_i,O^{h-1}_j,w_{ij}^h)$.
For all the output units, the local learning rules can be supervised since the targets are considered as local variables and thus of the form
$\Delta w^L_{ij}=F(T,O^L_i,O^{L-1}_j,w_{ij}^L)$.
}
\label{fig:DeepLocal}
\end{figure}

\par\null\par\noindent
{\bf Fact:}
Consider the supervised learning problem in a deep feedforward architecture with differentiable error function and transfer functions. Then in most cases deep local learning cannot find weights associated with critical points of the error functions, and thus it cannot find locally or globally optimal weights. 

\par\null\par\noindent
{\bf Proof:}
If we consider any weight $w^h_{ij}$ in a deep layer $h$ (i.e. $0<h < l$), a simple application of the chain rule (or the backpropagation equations) shows that

\be
\frac{\partial E_{err}}{\partial w^h_{ij}}=E \left [B^h_i(t)O^{h-1}_j(t) \right ]=\frac{1}{M}
\sum_{t=1}^M B^h_i(t) O^{h-1}_j(t)
\label{eq:th1}
\ee
where $B^h_i(t)$ is the backpropagated error of unit $i$ in layer $h$, which depends in particular on the targets $T(t)$ and the weights in the layers above layer $h$. Likewise, $O^{h-1}_j(t)$ is the presynaptic activity of unit $j$ in layer $h-1$ which depends on the inputs $I(t)$ and the weights in the layers below layer $h-1$. {\it In short, the gradient is a sum over all training examples of 
product terms, each product term being the product of a target-dependent term with an input-dependent term.} [{\it The target-dependent term depends explicitly also on all the descendant weights of unit $i$ in layer $h$, and the input-dependent term depends also on all the ancestors weights of unit $j$ in layer $h-1$}.]
As a result, in most cases, the deep weights $w^h_{ij}$, which correspond to a critical point where
$\partial E_{err}/{\partial w^h_{ij}}=0$, must 
depend {\it on both the inputs and the targets}, as well as all the other weights. In particular, this must be true at any local or global optimum. However, using any strictly local learning scheme all the deep weights $w_{ij}^h$ ($h<L$) depend on the {\it inputs only}, and thus cannot correspond to a critical point. 

In particular, this shows that applying local Hebbian learning to a feedforward architecture,
whether a simple autoencoder architecture or Fukushima's complex neocognitron architecture, cannot achieve optimal weights, regardless of which kind of local Hebbian rule is being  used. For the same reasons, an architecture consisting of a stack of autoencoders trained using unlabeled data only 
\cite{hinton2006fast,hinton2006reducing,bengio-lecun-07,bengio2007greedy,Erhan+al-2010}
cannot be optimal in general, even when the top layer is trained by gradient descent. It is of course possible to use local learning, shallow or deep autoencoders, Restricted Boltzmann Machines, and so forth to compress data, or to {it initialize} the weights of a deep architecture. However, these steps alone cannot learn complex functions optimally because learning a complex function optimally necessitates the reverse propagation of information from the targets back to the deep layers. 

The Fact above is correct at a level that would satisfy a physicist and is consistent with empirical evidence. It is not completely tight from a mathematical standpoint due to the phrase ``in most cases''. This expression is meant to exclude trivial cases that are not important in practice, but which would be difficult to capture exhaustively with mathematical precision.
These include the case when the training data is trivial with respect to the architecture (e.g. $M=1$) and can be loaded entirely in the weights of the top layer, even with random weights in the lower layers, or when the data 
is generated precisely with an artificially constructed architecture where the deep weights depend only on the input data, or are selected at random.

This simple result has significant consequences. In particular, if a constrained feedforward architecture is to be trained on a complex task in some optimal way, the deep weights of the architecture must depend on both the training inputs and the target outputs. Thus in any physical implementation, in order to be able to reach a locally optimal architecture there {\it  must exist a physical learning channel that conveys information about the targets back to the deep weights}. This raises three sets of questions regarding: (1) the nature of the backward learning channel; and (2) the nature of the information being transmitted through this channel; and (3) the rate of the backward learning channel. These questions will be addressed in Section \ref{sec:Channel}.
We now focus on the information about the targets that is being transmitted to the deep layers.

\section{Local Deep Learning and Deep Targets Algorithms}
\label{sec:DeepTargets}

\subsection{Definitions and their Equivalence}

We have seen in the previous section that in general in an optimal implementation each weight $w_{ij}^h$ must depend on both the inputs $I$ and the targets $T$. In order for learning to remain local, we let $I_{ij}^h(T)$ denote the information about the targets that is transmitted from the output layer to the weight $w_{ij}^h$ for its update by a corresponding local learning rule of the form

\be 
\Delta w_{ij}^h = F(I_{ij}^h,O^h_i,O^{h-1}_j,w_{ij}^{h})
\label{eq:}
\ee[The upper and lower indexes on $I$ distinguish it clearly from the inputs in the 0-th layer.
We call this local deep learning (Figure \ref{fig:LocalDeep}) in contrast with deep local learning. The main point of the previous section was to show that local deep learning is more powerful than deep local learning, and local deep learning is necessary for reaching optimal weights.

\begin{figure}[h!]
\begin{center}
\includegraphics[width=0.9\columnwidth]{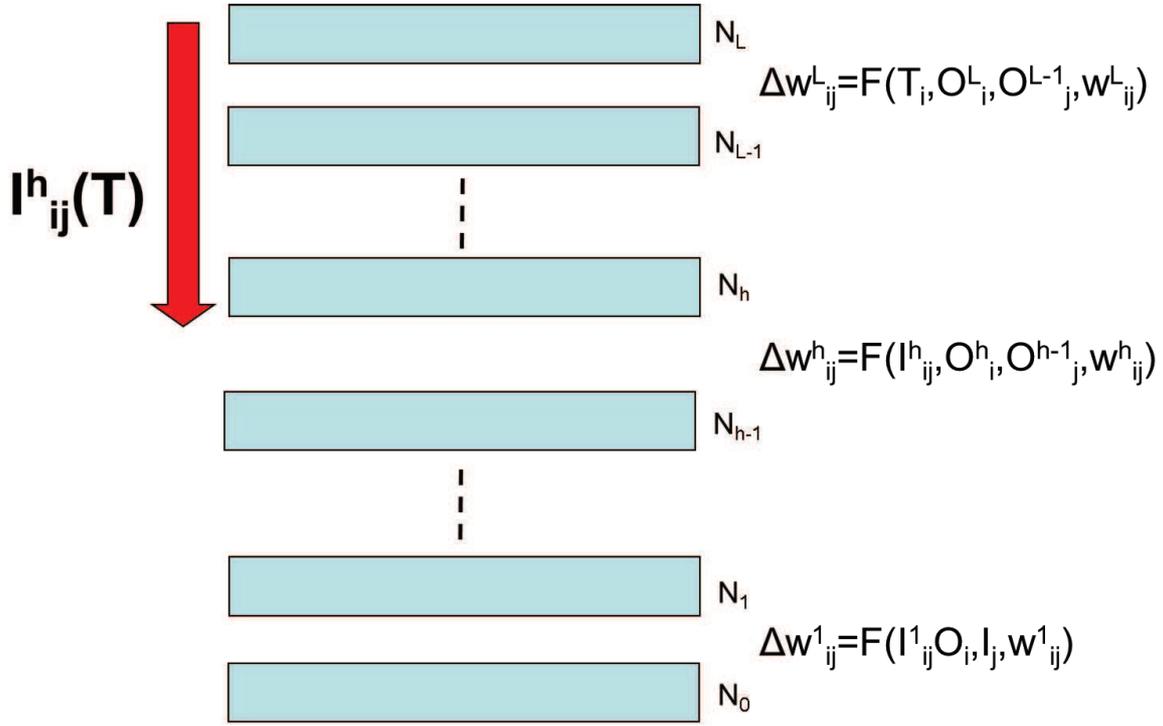}
\end{center}
\caption{\small Local deep learning. In general, deep local learning cannot learn complex functions optimally since it leads to architectures where only the weights in the top layer depend on the targets. For optimal learning, some information $I^h_{ij}(T)$ about the targets must be transmitted to each synapse associated with any deep layer $h$, so that it becomes a local variable that can be incorporated into the corresponding local learning rule $
\Delta w^h_{ij}=F(I_{ij}^h,O^h_i,O^{h-1}_j,w_{ij}^h)$.
 }
\label{fig:LocalDeep}
\end{figure}

\par \noindent
{\bf Definition 1:} Within the class of local deep learning algorithms, we define the subclass of deep targets local learning algorithms as those for which the information $I_{ij}^h$ transmitted about the targets depends only on the postsynaptic unit, in other words $I_{ij}^h(T)=I_i^h(T)$. Thus in a deep targets learning algorithm we have

\be 
\Delta w_{ij}^h = F(I_{i}^h,O^h_i,O^{h-1}_j,w_{ij}^{h})
\label{eq:}
\ee
for some function $F$ (Figure \ref{fig:DeepTargets}) .

We have also seen that when proper targets are available, there are efficient local learning rules for adapting the weights of a unit. In particular, the rule $\Delta w =\eta (T-O) I$ works well in practice for both sigmoidal and threshold transfer functions. Thus the deep learning problem can in principle be solved by providing good targets for the deep layers. We can introduce a second definition of deep targets algorithms:

\par
\null
\par\noindent
{\bf Definition 2:} A learning algorithm is a deep targets learning algorithm if it provides targets for all the trainable units.

\par\null\par
\noindent
{\bf Theorem:} Definition 1 is equivalent to Definition 2.
 Furthermore, backpropagation can be viewed as a deep targets algorithm.

\par\null\par
\noindent
{\bf Proof:}
Starting from Definition 2, if some target $T_i^h$ is available for unit $i$ in layer $h$, then we can set $I_i^h=T_i^h$ in Definition 1. Conversely, starting from Definition 1, consider a deep targets algorithm of the form

\be
\Delta w_{ij}^h = F(I_i^h,O_i^h,O_j^{h-1},w_{ij}^h)
\label{eq:deept1}
\ee
If we had a corresponding target $T_i^h$ for this unit, it would be able to learn by gradient descent in the form  

\be
\Delta w_{ij}^h =\eta (T_i^h-O_j^h)O_j^{h-1}
\label{eq:deept2}
\ee
This is true both for sigmoidal transfer functions and for threshold gates, otherwise the rule should be slightly modified to accommodate other transfer functions accordingly.
By combining Equations \ref{eq:deept1} and \ref{eq:deept2},
we can solve for the target 

\be 
T_i^h= \frac{ F(I_i^h,O_i^h,O_j^{h-1},w_{ij}^h)}{\eta O_j^{h-1}}+O_i^h
\label{eq:deept3}
\ee
assuming the presynaptic activity $O_j^{h-1} \neq 0$ (note that $T^L=T$) (Figure \ref{fig:DeepTargets}).
In particular, we see that backpropagation can be viewed as a deep targets algorithm providing targets for the hidden layers according to Equation \ref{eq:deept3}
in the form:

\be 
T_i^h = I^h_i + O^h_i
\label{eq:deept4}
\ee
where $I_i^h=B_i^h=\partial E_{err}/\partial S_I^h$ is exactly the backpropagated error.

\begin{figure}[h!]
\begin{center}
\includegraphics[width=0.9\columnwidth]{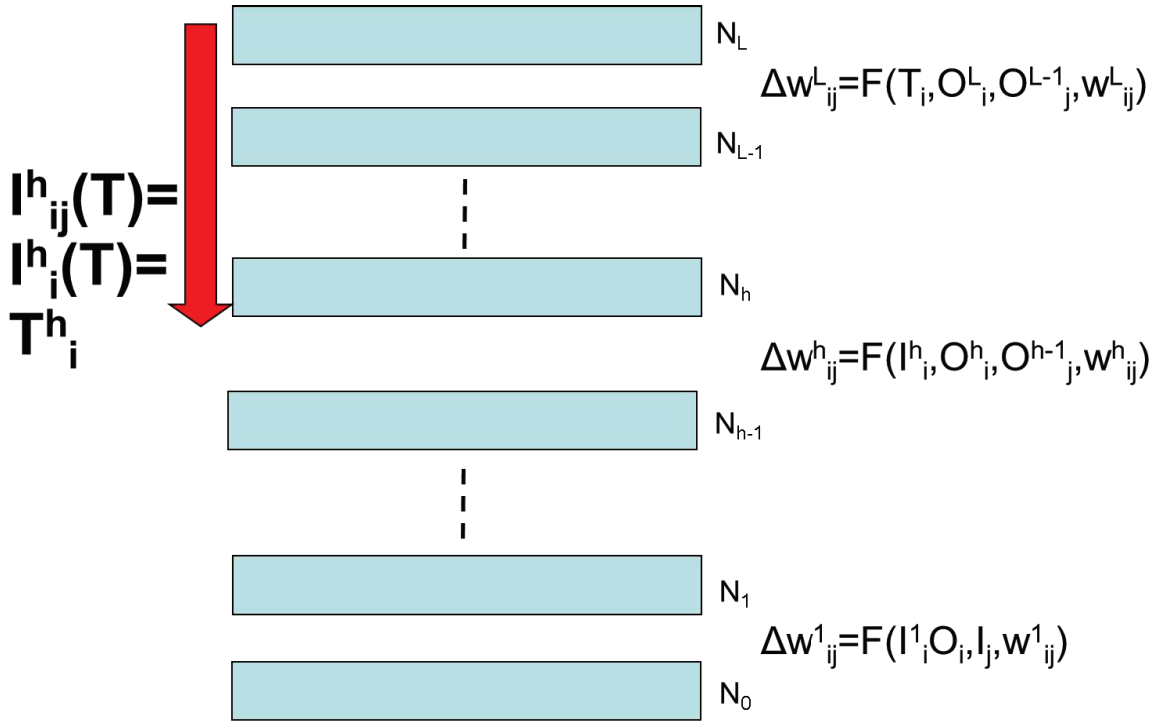}
\end{center}
\caption{\small Deep targets learning. This is a special case of local deep learning, where the transmitted information
$I^h_{ij}(T)$ about the targets does not depend on the presynaptic unit $I^h_{ij}(T)=I^h_i(T)$. It can be shown that this is equivalent to transmitting a deep target $T^h_i$ for training any unit $i$ in any deep layer $h$ by a local supervised rule of the form
$\Delta w^h_{ij}=F(T^h_i,O^h_i,O^{h-1}_j,w_{ij}^h)$. In typical cases (linear, threshold, or sigmoidal units), this rule is $\Delta w^h_{ij}=\eta(T^h_i - O^h_i)O^{h-1}_j$.
}
\label{fig:DeepTargets}
\end{figure}


\subsection{Deep Targets Algorithms: the Sampling Approach}

\begin{figure}[h!]
\begin{center}
\includegraphics[width=0.8\columnwidth]{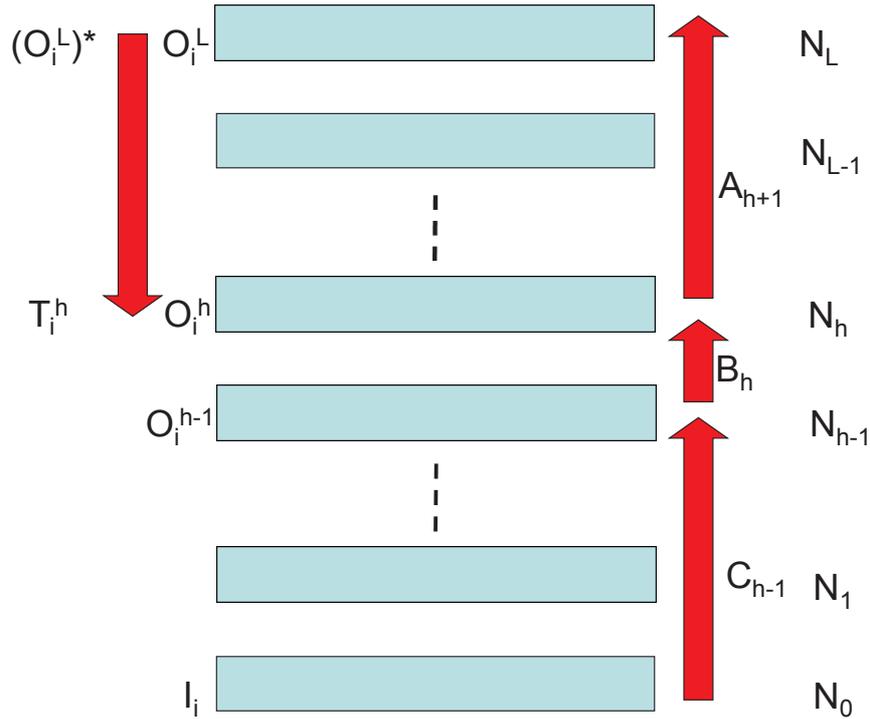}
\end{center}
\caption{\small Deep architecture and deep targets algorithm. The algorithm visits the various layers according to some schedule and optimizes each one of them. This is achieved by the deep targets algorithm which is capable of providing suitable targets $(T_i^h)$ for any layer $h$ and any input $I$, assuming that the rest of the architecture is fixed. The targets are used to modify the weights associated with the function $B_h$. The targets can be found using a sampling strategy: sampling the activities in layer $h$, propagating them forward to the output layer, selecting the best output sample, and backtracking it to  the best sample in layer $h$.}
\label{fig:DeepAlgorithm}
\end{figure}

In the search for alternative to backpropagation, one can thus investigate whether there exists alternative deep targets algorithms \cite{sadowskibaldinips12}.
More generally, deep targets algorithms rely on two key assumptions: (1) the availability of an algorithm 
$\Theta$ for optimizing any layer or unit, while holding the rest of the architecture fixed, once a target is provided; and (2) the availability of an algorithm for providing deep targets. The maximization by $\Theta$ may be complete or partial, this optimization taking place with respect to an error measure
$\Delta^h$ that can be specific to layer $h$ in the architecture (or even specific to a subset of units in the case of an architecture where different units are found in the same layer). For instance, an exact optimization algorithm $\Theta$ is obvious in the unconstrained Boolean case \cite{baldiboolean12,baldijmlr12}. For a layer of threshold gates, $\Theta$ can be the perceptron algorithm, which is exact in the linearly separable case. For a layer of artificial neurons with differentiable transfer functions, $\Theta$ can be the delta rule or gradient descent, which in general perform only partial optimization. Thus deep targets algorithms proceed according to two loops: an outer loop and an inner loop. The inner loop is used to find suitable targets. The outer loop uses these targets to optimize the weights, as it cycles through the units and the layers of the architecture.

\subsubsection{Outer Loop}

The outerloop is used to cycle through, and progressively modify, the weights of a deep feedforward architecture:
\begin{enumerate}
\item Cycle through the layers and possibly the individual units in each layer according to some schedule. Examples of relevant schedules include successively sweeping through the architecture layer by layer from the first layer to the top layer. 
\item During the cycling process, for a given layer or unit, identify suitable targets, while holding the rest of the architecture fixed.
\item Use the algorithm $\Theta$ to optimize the corresponding weights.
\end{enumerate}
Step 2 is addressed by the following inner loop.

\subsubsection{Inner Loop: the Sampling Approach}

The key question of course is whether one can find ways for identifying deep targets $T_i^h$, other than backpropagation, which is available only in differentiable networks. It is possible to identify targets by using a sampling strategy in both differentiable and non-differentiable networks.

In the online layered version, consider an input vector $I=I(t)$ and its target $T=T(t)$ and any adaptive layer $h$, with $ 1 \leq h \leq L$. We can write the overall input-output function $W$ as $W=A_{h+1}B_hC_{h-1}$ (Figure \ref{fig:DeepAlgorithm}). We assume that both $A_{h+1}$ and $C_{h-1}$ are fixed.
The input $I$ produces an activation vector $O_i^{h-1}$ and our goal is to find a suitable vector target $T^h$ for layer $h$.
For this we generate a sample ${\cal S}^h$ of activity vectors $S^h$ in layer $h$. This sampling can be carried in different ways, for instance: (1) by sampling the values
$O^h$ over the training set; (2) by small random perturbations, i.e. using random vectors sampled in the proximity of the vector $O^h=B_hC_{h-1}(I)$;(3) by large random perturbation (e.g. in the case of logistic transfer functions by tossing dies with probabilities equal to the activations) or by sampling uniformly; and (4) exhaustively (e.g. in the case of a short binary layer). 
Finally, each sample $S^h$ can be propagated forward to the output layer and produce a corresponding output $A_{h+1}(S^h)$. We then select as the target vector $T^h$ the sample that produces the output closest to the true target $T$. Thus

\be
T^h= \arg \min_{S^h\in {\cal S}^h} \Delta^L(T,A_{h+1}(S^h))
\label{eq:}
\ee
If there are several optimal vectors in ${\cal S}^hj$, then one can select one of them at random, or use $\Delta^h$ to control the size of the learning step. For instance, by selecting a vector $S^h$ that not only minimizes the output error $\Delta^L$ but also minimizes the error
$\Delta^h(S^h,O^h)$, one can ensure that the target vector is as close as possible to the current layer activity, and hence minimizes the corresponding perturbation. As with other learning and optimization algorithms, these algorithmic details can be varied during training, for instance by progressively reducing the size of the learning steps as learning progresses
(see Appendix D for additional remarks on deep targets algorithms). Note that the algorithm described above is  the natural generalization of the algorithm  introduced in \cite{baldiboolean12} for the unrestricted Boolean autoencoder, specifically for the optimization of the lower layer. Related reinforcement learning
\cite{sutton1998reinforcement} algorithms for connectionist networks of stochastic units can be found in \cite{williams1992simple}.]

\subsection{Simulation}

Here we present the result of a simulation to show that sampling deep target algorithms can work and can even be applied to the case of non-differentiable networks where back propagation cannot be applied directly. A different application of the deep targets idea is developed in \cite{deepcontact2012}.
We use a four-adjustable-layer perceptron autoencoder with threshold gate units and Hamming distance error in all the layers. The input and output layers have $N_0=N_4=100$ units each, and there are three hidden layers with $N_1=30$, $N_2=10$, and $N_3=30$ units. All units in any layer $h$ are fully connected to the $N_{h-1}$ units in the layer below, plus a bias term. The weights are initialized randomly from the uniform distribution $U(-\frac{1}{\sqrt{N_{h-1}}}, \frac{1}{\sqrt{N_{h-1}}})$ except for the bias terms which are all zero.

The training data consists of 10 clusters of 100 binary examples each for a total of $M=1000$. The centroid of each cluster is a random 100-bit binary vector with each bit drawn independently from the binomial distribution with $p=0.5$. An example from a particular cluster is generated by starting from the centroid and introducing noise -- each bit has an independent probability $0.05$ of being flipped. The test data consists of an additional 100 examples drawn from each of the 10 clusters.
The distortion function $\Delta^h$ for all layers is  the Hamming distance, and the optimization algorithm $\Theta$ is 10 iterations of the perceptron algorithm with a learning rate of $1$. The gradient is calculated in batch mode using all $1000$ training examples at once. 
For the second layer with $N_2 = 10$, we use exhaustive sampling since there are only $2^{10}=1024$ possible activation values.
For other layers where $N_h > 10$, the sample comprises all the 1000 activation vectors of the corresponding layer over the training set, plus a set of $1000$ random binary vectors where each bit is independent and $1$ with probability $0.5$.
Updates to the layers are made on a schedule that cycles through the layers in sequential order: $1,2,3,4$. One cycle of updates constitutes an epoch. 
The trajectory of the training and test errors are shown in Figure \ref{fig:thresholdgate_auto} demonstrating that this sampling deep targets algorithm is capable of training this non-differentiable network reasonably well.

\begin{figure}[h!]
\begin{center}
\includegraphics[width=0.8\columnwidth]{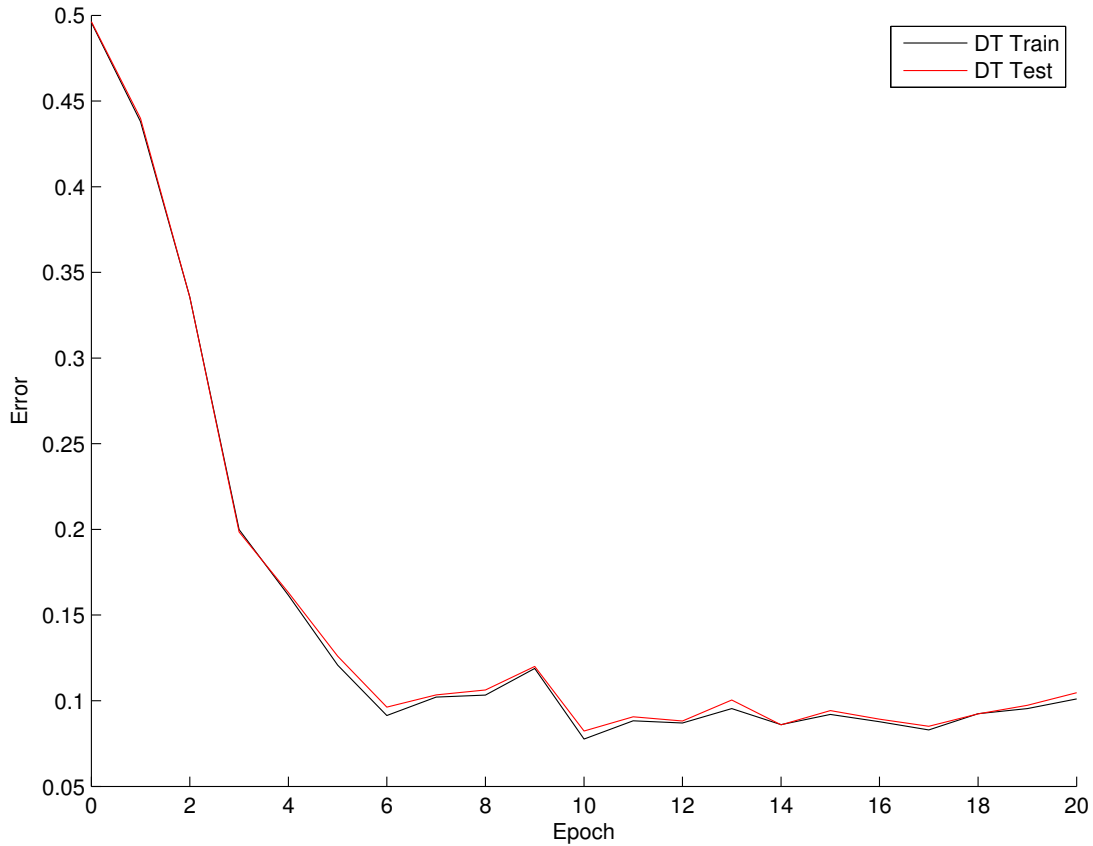}
\end{center}
\caption{\small A sampling deep target (DT) algorithm is used to train a simple autoencoder network of threshold gates thus purely comprised of non-differentiable transfer functions. The $y$ axis correspond to the average Hamming error per component and per example.}
\label{fig:thresholdgate_auto}
\end{figure}

\section{The Learning Channel and the Optimality of Backpropagation}
\label{sec:Channel}
Armed with the understanding that in order to implement learning capable of reaching minima of the error function there must be a channel conveying information about the targets to the deep weights, we can now examine the three key questions about the channel:  its nature, its semantics, and its rate.

\subsection{The Nature of the Channel}

In terms of the nature of the channel, regardless of the hardware embodiment, there are two main possibilities. Information about the targets can travel to the deep weights either by: (1) traveling along the physical forward connections but in the reverse direction; or (2) using a separate different channel.

\subsubsection{Using the Forward Channel in the Backward Direction}

In essence, this is the implementation that is typically emulated on digital computers using the transpose of the forward matrices in the backpropagation algorithm. However,
the first thing to observe, is that even when the same channel is being used in the forward and backward direction, the signal itself does not need to be of the same nature.  For instance the forward propagation could be electrical while the backward propagation could be chemical. This is congruent with the observation that the two signals can have very different time scales with the forward propagation being fast compared to learning which can occur over longer time scales. In biological neural networks, there is evidence for the existence of complex molecular signaling cascades traveling from the synapses of a neuron to the DNA in its nucleus,  capable of activating epigenetics modifications and gene expression, and conversely for molecular signals traveling from the DNA to the synapses. At least in principle, chemical or electrical signals could traverse synaptic clefts in both directions. In short,
while there is no direct evidence supporting the use of the same physical connections in both directions in biological neural systems, this possibility cannot be ruled out entirely at this time and conceivably it could be used in other hardware embodiments. It should be noted that a deep targets algorithm in which the feedback reaches the soma of a unit leads to a simpler feedback channel but puts the burden on the unit to propagate the central message from the soma to the synapses.

\subsubsection{Using a Separate Backward Learning Channel}

If a separate channel is used, the channel must allow the transfer of information from the output layer and the targets to the deep weights. This transfer of information could occur through direct connections from the output layer to each deep layer, or through a staged process of propagation through each level (as in backpropagation). Obviously combination of both processes are also possible. In either case, the new channel implements some form of feedback.  Note again that the learning feedback can be slow and distinct in terms of signal, if not also in terms of channel, from the usual feedback of recurrent networks that is typically used in rapid dynamic mode to fine tune a rapid response, e.g. by helping combine a top down generative model with a bottom up recognition model.

In biological neuronal circuits, there are plenty of feedback connections between different processing stages (e.g. \cite{felleman1991distributed}) and some of these connections could serve as the primary channel for carrying the feedback signal necessary for learning. It must be noted, however, that given a synaptic weight $w_{ij}$, the feedback could typically reach either the dendrites of neuron $i$, or the dendrites of neuron $j$ or both. In general, the dendrites of the presynaptic neuron $j$ are physically far away from the synapse associated with $w_{ij}$ which is located on the dendritic tree of the post-synaptic neuron $i$, raising again a problem of information transmission within neuron $j$ from its incoming synapses on its dendritic tree to the synapses associated with its output at the end of the arborization of its axon. The feedback reaching the dendrites of neuron $i$ could in principle be much closer to the site where $w_{ij}$ is implemented, although this requires a substantial degree of organization of the synapses along the dendrites of neuron $i$, spatially combining synapses originating from feedforward neurons with synapses originating from feedback neurons, together with the biochemical mechanisms required for the local transmission of the feedback information between spatially close synapses.
In short, the nature of the feedback channel depends on the physical implementation. While a clear understanding of how biological neural systems implement learning is still out of reach, the framework presented here--in particular the notions of local learning and deep targets algorithms--clarifies some of the aspects and potential challenges associated with the feedback channel and the complex geometry of neurons.

An important related issue is the issue of the symmetry of the weights. Backpropagation uses symmetric (transposed) weights in the forward and backward directions in order to compute exact gradients. In a physical implementation, especially one that uses different channels for the forward and backward propagation of information, it may be difficult to instantiate weights that are precisely symmetric.
However simulations \cite{lillicrap2014random} seem to indicate that, for instance, random weights can be used in the backward direction without affecting too much the speed of learning, or the quality of the solutions.
Random weights in general result in matrices that have the maximum rank allowed by the size of the layers and thus transmit as much information as possible in the reverse direction, at least globally at the level of entire layers. How this global transmission of information allow precise learning is not entirely clear. But at least in the simple case of a network with one hidden layer and one output unit, it is easy to give a mathematical proof that random weights will support convergence and learning, provided the random weights have the same sign as the forward weights. It is plausible that biological networks could use non-symmetric connections, and that these connections could possibly be random, or random but with the same sign as the forward connections.

\subsection{The Semantics of the Channel}

Regardless of the nature of the channel, next one must consider the meaning of the information that is being transmitted to the deep weights, as well as its amount. Whatever information about the targets is fed back, it is ultimately used within each epoch to change the weights in the form 

\be
\Delta w_{ij}^h =\eta_{ij}^h E[F(I_{ij}^h,O_i^h,O_j^{h-1},w_{ij}^h)]
\label{eq:}
\ee
so that with small learning rates $\eta_{ij}^h$ a Taylor expansion leads to

\begin{eqnarray} 
E_{err}(w_{ij}^h+\Delta w_{ij}^h) &=&
E_{err}(w_{ij}^h)+\sum_{w_{ij^h}} \frac{\partial E_{err}}{\partial w_{ij}^h} \Delta w_{ij}^h +\frac{1}{2} (\Delta w_{ij}^h)^t H
(\Delta w_{ij}^h) +R \\
&=&E_{err}(w_{ij}^h)+ G \cdot (\Delta w_{ij}^h)
+\frac{1}{2} (\Delta w_{ij}^h)^t \cdot H
(\Delta w_{ij}^h) +R
\label{eq:deept5}
\end{eqnarray}
where $G$ is the gradient, $H$ is the Hessian, and $R$ is the higher order remainder. If we let $W$ denote the total number of weights in the system, the full Hessian has $W^2$ entries and thus in general is not computable for large $W$, which is the case of interest here. Thus limiting the expansion to the first order:

\be
E_{err}(w_{ij}^h+\Delta w_{ij}^h) \approx 
E_{err}(w_{ij}^h) + G \cdot(\Delta w_{ij}^h)=
E_{err}(w_{ij}^h) + \eta \vert \vert G \vert \vert 
u \cdot g=E_{err}(w_{ij}^h) + \eta \vert \vert G \vert \vert {\cal O}
\label{eq:}
\ee
where $u$ is the unit vector associated with the weight adjustments ($\eta u= (\Delta w_{ij}^h$)), $g$ is the unit vector associated with the gradient ($g=G/\vert\vert G\vert\vert$), and ${\cal {O}}=g \cdot u$. Thus to a first order approximation, the information that is sent back to the deep weights can be interpreted in terms of how well it approximates the gradient $G$, or how many bits of information it provides about the gradient. With $W$ weights and a precision level of $D$-bits for any real number, the gradient contains $WD$ bits of information. These can in turn be split into $D-1$ bits for the magnitude $\vert \vert G \vert \vert$ of the gradient (a single positive real number), and 
$(W-1)D+1$ bits to specify the direction by the  corresponding unit vector ($g=G/\vert\vert G \vert \vert$), using $W-1$ real numbers
plus one bit to determine the sign of the remaining component.
Thus most of the information of the gradient in a high-dimensional space is contained in its {\it direction}. The information $I^h_{ij}$ determines $\Delta w_{ij}^h$ and thus the main questions is how many bits the vector $(\Delta w_{ij}^h)$ conveys about $G$, which is essentially how many bits the vector $u$
conveys about $g$, or how close is $u$ to $g$? With a full budget of $B$ bits per weight, the gradient can be computed with $B$ bits of precision, which defines a box around the true gradient vector, or a cone of possible unitary directions $u$. Thus the expectation of $\cal O$ provides a measure of how well the gradient is being approximated and how good is the corresponding optimization step (see next section).

Conceivably, one can also look at regimes where even {\it more} information than the gradient is transmitted through the backward channel. This could include, for instance, second order information about the curvature of the error function. However, as mentioned above, in the case of large deep networks this seems problematic since with $W$ weights, this procedure would scale like $W^2$. Approximations that essentially compute only the diagonal of the Hessian matrix,
and thus only $W$ additional numbers, have been considered \cite{LeCun:90a}, using a procedure similar to backpropagation that scales like $W$ operations. These methods were introduced for other purposes (e.g. network pruning) and do not seem to have led to significant or practically useful improvements to deep learning methods. Furthermore, they do not change the essence of following scaling computations.

\subsection{The Rate and other Properties of the Channel}

Here we want to compare several on-line learning algorithms where information about the targets is transmitted back to the deep weights and define and compute a notion of transmission rate for the backward channel. We are interested in estimating a number of important quantities in the limit of large networks. The estimates do not need to be very precise, we are primarily interested in expectations and scaling behavior. Here all the estimates are computed for the adjustment of all the weights on a given training example and thus would have to be multiplied by a factor $M$ for a complete epoch. In particular, given a training example, we want to estimate the scaling of:

\begin{itemize}
\item The number $\cal C_W$ of computations required to transmit the backward information per network weight. The estimates are computed in terms of number of elementary operations which are assumed to have a fixed unit cost. Elementary operations include addition, multiplication, computing the value of a transfer function, and computing the value of the derivative of the transfer function.
We also assume the same costs for the forward or backward propagation of information. Obviously these assumptions in essence capture the implementation of neural networks on digital computers but could be revised when considering completely different physical implementations. With these assumptions, the total number of computations required by a forward pass or a backpropagation through the network scales like $W$, and thus ${\cal C_W}=1$ for a forward or backward pass.

\item The amount of information $\cal I_W$ that is sent back to each weight. In the case of deep targets algorithms, we can also consider the amount of information $\cal I_N$ that is sent back to each hidden unit, from which a value $\cal I_W$ can also be derived. We let $D$ (for double precision) denote the number of bits used to represent a real number in a given implementation. Thus, for instance, the backpropagation algorithm provides $D$ bits of information to each unit and each weight (${\cal I_W}={\cal I_N}=D$) for each training example, associated with the corresponding derivative.

\item We define the {\it rate} $\cal R $ of the backward channel of a learning algorithm by ${\cal R}={\cal I_W}/{\cal C_W}$. It is the number of bits (about the gradient) transmitted to each weight through the backward channel divided by the number of operations  required to compute/transmit this information per weight. Note that the rate is bounded by $D$:  ${\cal R} \leq D$. This is because the maximal information that can be transmitted is the actual gradient corresponding to $D$ bits per weight, and the minimal computational/transmission cost must be at least one operation per weight.

\item It is also useful to consider the improvement or expected improvement $\cal O'$, or its normalized version $\cal O$. All the algorithms to be considered ultimately lead to a learning step $\eta u$ where $\eta$ is the global learning rate and $u$ is the vector of weight changes. To a first order of approximation, the corresponding improvement is computed by taking the dot product with the gradient so that ${\cal { O'}}=\eta u \cdot  G$. In the case of (stochastic) gradient descent we have ${\cal O'}=\eta G \cdot  G=\eta \vert\vert G \vert \vert^2$. In gradient descent, the gradient provides both a direction and a magnitude of the corresponding optimization step. 
In the perturbation algorithms to be described, the perturbation stochastically produces a direction but there is no natural notion of magnitude. Since when $W$ is large most of the information about the gradient is in its direction (and not its magnitude), to compare the various algorithms we can simply compare the directions of the vector being produced,in particular in relation to the direction of the gradient. Thus we will assume that all the algorithms produce a step of the form $\eta u$
where $\vert \vert u \vert \vert =1$ and thus 
${\cal O'}= \eta u \cdot  G=\eta \vert \vert G \vert \vert u \cdot  g=\eta \vert \vert G \vert \vert {\cal O}$. Note that the maximum possible value of ${\cal O}=u \cdot g$ is one, and corresponds to ${\cal {O'}}=\eta \vert \vert G \vert \vert$.
\end{itemize}

To avoid unnecessary mathematical complications associated with the generation of random vectors of unit length uniformly distributed over a high-dimensional sphere, we will approximate this process by assuming in some of the calculations that the components of $u$ are i.i.d. Gaussian with mean 0 and variance $1/W$ (when perturbing the weights).
Equivalently for our purposes, we can alternatively  assume the components of $u$ to be i.i.d. uniform over $[-a,a]$, with $a={\sqrt {3/W}}$, so that the mean is also $0$ and variance $1/W$. In either case, the square of the norm $u^2$ tends to be normally distributed by the central limit theorem, with expectation 1. A simple calculation shows that the variance of $u^2$ is given by $2/W$ in the Gaussian case, and by $4/5W$ in the uniform case.
Thus in this case $G \cdot  u$ tends to be normally distributed with mean 0 and variance $C\vert\vert G \vert \vert^2/W $ (for some constant $C>0$) so that
${\cal O'} \approx \eta {\sqrt C}  \vert \vert G \vert\vert/{\sqrt {W}} $.

In some calculations, we will also require all the components of $u$ to be positive. In this case, it is easier to assume that the 
components of $u$ are i.i.d. uniform over $[0,a]$ with 
$a={\sqrt{3/W}}$, so that $u^2$ tends to be normally distributed by the central limit theorem, with expectation 1 and variance
$4/5W$.
Thus in this case $G \cdot  u$ tends to be normally distributed with mean $({\sqrt{3/W}/2}) \sum_iG_i $ and variance
$    \vert \vert G \vert \vert^2/(4W) $  so that
${\cal O'} \approx \eta ({\sqrt{3/W}/2}) \sum_iG_i  $.

\subsection{A Spectrum of Descent Algorithms}

In addition to backpropagation (BP) which is one way of implementing stochastic gradient descent, we consider stochastic descent algorithms associated with small perturbations of the weights or the activities. These algorithms can be identified by a name of the form 
P\{W or A\}\{L or G\}\{B or R\}\{$\emptyset$ or K\}. The perturbation (P) can be applied to the weights (W) or, in deep targets algorithms, to the activities (A).
The perturbation can be either local (L) when applied to a single weight or activity, or global (G) when applied to all the weights or activities. The feedback provided to the network can be either binary (B) indicating whether the perturbation leads to an improvement or not, or real (R) indicating the magnitude of the improvement. Finally, the presence of K indicates that the corresponding  perturbation is repeated $K$ times. 
For brevity, we focus on the following main cases (other cases, including intermediary cases between local and global where, for instance, perturbations are applied layerwise can be analyzed in similar ways and do not offer additional insights or improvements): 
 
\begin{itemize}
 \item  PWGB is the stochastic descent algorithm where all the weights are perturbed by a small amount. If the error decreases the perturbation is accepted. If the error increases the perturbation is rejected. Alternatively the opposite perturbation can be accepted, since it will decrease the error (in the case of differentiable error function and small perturbations), however this is detail since at best it speeds things up by a factor of two.
 \item PWLR is the stochastic descent algorithm where each weight in turn is perturbed by a small amount and the feedback provided is a real number representing the change in the error. Thus this algorithm corresponds to the computation of the derivative of the error with respect to each weight using the definition of the derivative. In short, it corresponds also to stochastic gradient descent but provides a different mechanism for computing the derivative. It is not a deep targets algorithm.
 \item PWLB is the binary version of PWLR where only one bit, whether the error increases or decreases, is transmitted back to each weight upon its small perturbation. Thus in essence this algorithms provides the sign of each component of the gradient, or the orthant in which the gradient is located, but not its magnitude. After cycling once through all the weights, a random descent unit vector can be generated in the corresponding orthant (each component of $u_i$ has the sign of $g_i$).
 \item PALR is the deep targets version of PWLR where the activity of each unit in turn is perturbed by a small amount, thus providing the derivative of the error with respect to each activity, which in turn can be used to compute the derivative of the error with respect to each weight.
  \item    PWGBK is similar to PWGB, except that $K$ small global perturbations are produced, rather than a single one. In this case, the binary feedback provides information about which perturbation leads to the largest decrease in error.
 \item  PWGRK is similar to PWGBK except that for each perturbation a real number, corresponding to the change in the error, is fed back. This corresponds to providing the value of the dot product of the gradient with $ K$ different unit vector directions.
\end{itemize}

\subsection{Analysis of the Algorithms: the Optimality of Backpropagation}

\subsubsection{Global Weight Perturbation with Binary Feeback (PWGB)}

For each small global perturbation of the weights, this algorithm transmits a single bit back to all the weights, corresponding to whether the error increases or decreases. This is not a deep targets algorithm. The perturbation itself requires one forward propagation, leading to ${\cal I_W}=1/W$ and ${\cal C_W}=1$. Thus:

\begin{itemize}
\item  ${\cal {I_W}}=1/W$
\item  ${\cal {C_W}}=1$
\item ${\cal { R}}=1/W$
\item  ${\cal {O'}}=\eta {{C}} 
\vert \vert G \vert\vert / \sqrt W$ for some constant $C>0$,
so that ${\cal O}=C/\sqrt W$
\end{itemize}

\subsubsection{Local Weight Perturbation with Real Feedback (PWLR)}

This is the definition of the derivative.
The derivative $\partial E_{err}/\partial w_{ij}^h$ can also be computed directly by first perturbing $w_{ij}^h$ by a small amount $\epsilon$, propagating forward, measuring $\Delta E_{err}=E_{err}(w_{ij}+\epsilon)-E_{err}(w_{ij})$ and then using 
$\partial E_{err}/\partial w_{ij}\approx \Delta E_{err}/\epsilon$. This is not a deep target algorithm. The algorithm computes the gradient and thus propagates $D$ bits back to each weight, at a total computational cost that scales like $W$ per weight, since it essentially requires one forward propagation for each weight. Thus:

\begin{itemize}
\item  ${\cal {I_W}}=D$
\item  ${\cal{ C_W}}=W$
\item ${\cal{ R}}=D/W$
\item ${\cal {O'}}= \eta \vert \vert G \vert\vert   $ for a step $\eta g$, and thus  ${\cal O}=1$
\end{itemize}

\subsubsection{Local Weight Perturbation with Binary Feedback (PWLB)}

 This is not a deep target algorithm. The algorithm provides a single bit of information back to each weigh and requires a forward propagation to do so. Without any loss of generality, we can assume that all the components of the final random descent vector $u_i$ must be positive.  Thus:

\begin{itemize}
\item  ${\cal {I_W}}=1$
\item  ${\cal{ C_W}}=W$
\item ${\cal{ R}}=1/W$
\item ${\cal {O'}}= \eta ({\sqrt{3/W}/2}) \sum_iG_i $, and thus   
 ${\cal O}=({\sqrt{3/W}/2}) \sum_ig_i$
\end{itemize}

\subsubsection{Local Activity Perturbation with Real Feedback (PALR)}

This is a deep target algorithm and from the computation of the derivative with respect to the activity of each unit, one can derive the gradient. So it provides $D$ bits of feedback to each unit, as well as to each weight. The algorithm requires in total $N$ forward propagations, one for each unit, resulting in a total computational cost of $NW$ or $N$ per weight. Thus:

\begin{itemize}
\item  ${\cal {I_W}}= {\cal {I_N}}=D$
\item  ${\cal {C_W}}=N$
\item ${\cal {R}}=D/N$
\item ${\cal {O'}}=\eta \vert \vert G \vert\vert $ for a step $\eta g$, and thus ${\cal O}=1$
\end{itemize}

\subsubsection{Global Weight Perturbation with Binary Feedback $K$ Times (PWGBK)}

In this version of the algorithm, the information backpropagated is which of the $K$ perturbation leads to the best improvement, corresponding to the same $\log K$ bits for all the weights. The total cost is $K$ forward propagations. Note that the $K$ perturbations constrain the gradient to be in the intersection of $K$ hyperplanes, and this corresponds to more information than 
retaining only the best perturbation.
However the gain is small enough that a more refined version of the algorithm and the corresponding calculations are not worth the effort. Thus here we just use the best perturbation. We have seen that for each perturbation the dot product of the corresponding unit vector $u$ with $G$ is essentially normally distributed with mean 0 and variance  $C \vert \vert G \vert \vert^2/W$. The maximum of $K$ samples of a normal distribution (or the absolute value of the samples if random ascending directions are inverted into descending directions) follows an extreme value distribution 
\cite{coles01,galambos87}
and the average of the maximum will scale like the standard deviation times a factor ${\sqrt {\log K}}$  up to a multiplicative constant. 
Thus:

\begin{itemize}
\item  ${\cal {I_W}}=\log K/W$
\item  ${\cal {C_W}}=K$
\item ${\cal {R}}=(\log K /W) /K$
\item ${\cal {O'}}=\eta { {C}} 
\vert \vert G \vert\vert {\sqrt {\log K}} / \sqrt W$ for some constant $C>0$, and thus ${\cal O}=C {\sqrt {\log K}} / \sqrt W$
\end{itemize}

\subsubsection{Global Weight Perturbation with Real Feedback $K$ Times (PWGRK)}

This algorithm provides $KD$ bits of feedback in total, or $KD/W$ per weight and requires $K$ forward propagations. In terms of improvements, let us consider that the algorithms generates $K$ random unit vector directions $u^{(1)},\ldots,u^{(K)}$ and produces the $K$ dot products $u^{(1)}\cdot G,\ldots, u^{(K)}\cdot G$.
In high dimensions ($W$ large), the $K$ random directions are approximately orthogonal
As a result of this information, one can select the unit descent direction given by

\be 
u=\frac{\sum_{k=1}^K  (u^{(k)}\cdot G)u^{(k)}}
{\vert \vert \sum_{k=1}^K  (u^{(k)}\cdot G)u^{(k)}\vert \vert }
\label{eq:}
\ee
Now we have

\be 
\vert \vert \sum_{k=1}^K  (u^{(k)}\cdot G)u^{(k)}\vert \vert ^2 \approx  \sum_{k=1}^K (u^{(k)} \cdot G)^2\approx CK \frac{\vert \vert G \vert \vert^2} {W}
\label{eq:}
\ee
for some constant $C>0$. The first approximation is because the vectors $u^{(k)}$ are roughly orthogonal, and the second is simply by taking the expectation.
As a result, ${\cal {O}}=\eta u \cdot G = 
\eta C {\sqrt {K}} \vert \vert G \vert \vert/{\sqrt{W}}$
for some constant $C>0$. Thus:

\begin{itemize}
\item  ${\cal {I_W}}=KD/W$
\item  ${\cal {C_W}}=K$
\item ${\cal {R}}=D/W$
\item ${\cal {O'}}=  \eta C {\sqrt {K}} \vert \vert G \vert \vert/{\sqrt{W}}$, and thus ${\cal O}=C {\sqrt {K}}/  {\sqrt{W}}$
\end{itemize}

\begin{table}[h!]
\renewcommand{\arraystretch}{1.5}
\begin{center}
    \begin{tabular}{ | c | l|l|l | l|  }
    \hline
 \textbf{Algorithm} & \textbf{Information $\cal I_W$}&
 \textbf{Computation $\cal C_W$}&
  \textbf{Rate $\cal R$} &\textbf{Improvement $\cal O$}  \\ \hline
 PWGB &$1/W$&1&$1/W$  & $C/\sqrt W$\\ \hline
  PWLR &$D$&$W$& $D/W$   & 1 \\ \hline
 PWLB &1&$W$&$1/W$ &   $ (\sqrt {3/W}/2)\sum_ig_i$ \\\hline
 PALR   &$D$&$N$&$D/N$ & 1 \\ \hline
 PWGBK &$\log K/W$    &K&$(\log K /W) /K$ & $C\sqrt{\log K} /\sqrt W$ \\ \hline
 PWGRK &$KD/W$&$K$&$D/W$ & $C\sqrt K/\sqrt W$\\ \hline
 BP   &D&1 & $D$ & 1 \\\hline 
    \end{tabular}
\end{center}
\caption{The rate $\cal R$ and improvement $\cal O$ of several optimization algorithms.}
 \label{tab:algoprop}
\end{table}

\subsubsection{Backpropagation (BP)}

As we have already seen:

\begin{itemize}
\item ${\cal {I_W}}={\cal {I_N}}=D$ 
\item ${\cal {C}}=1$
\item ${\cal {R}}=D$
\item ${\cal {O'}}=\eta \vert \vert G \vert\vert $ for a step $\eta g$, and thus ${\cal O}=1$
\end{itemize}

The reason that no algorithms better than backpropagation has been found is that the rate $\cal R$ of backpropagation is greater or equal to that of all the alternatives considered here
(Table \ref{tab:algoprop}). This is true also for the improvement $\cal O$. 
Furthermore, there is no close second: all the other algorithms discussed in this section fall considerably behind backpropagation in at least one dimension.
And finally, it is unlikely that an algorithm exists with a rate or improvement higher than backpropagation, because backpropagation achieves both the maximal possible rate, and maximal possible improvement (Figure \ref{fig:algos}), up to multiplicative constants.
Thus in conclusion we have the following theorem:
\par
\null\par 
\noindent
{\bf Theorem:}
The rate $\cal R$ of backpropagation is above or equal to the rate of all the other algorithms described here and it achieves the maximum possible value ${\cal R}=D$. 
The improvement $\cal O$ of backpropagation is above or equal to the improvement of all the other algorithms described here and it achieves the maximum possible value ${\cal {O}}=1$ (or ${\cal O'}=\eta \vert \vert G \vert \vert$).

\begin{figure}[h!]
\begin{center}
\includegraphics[width=0.8\columnwidth]{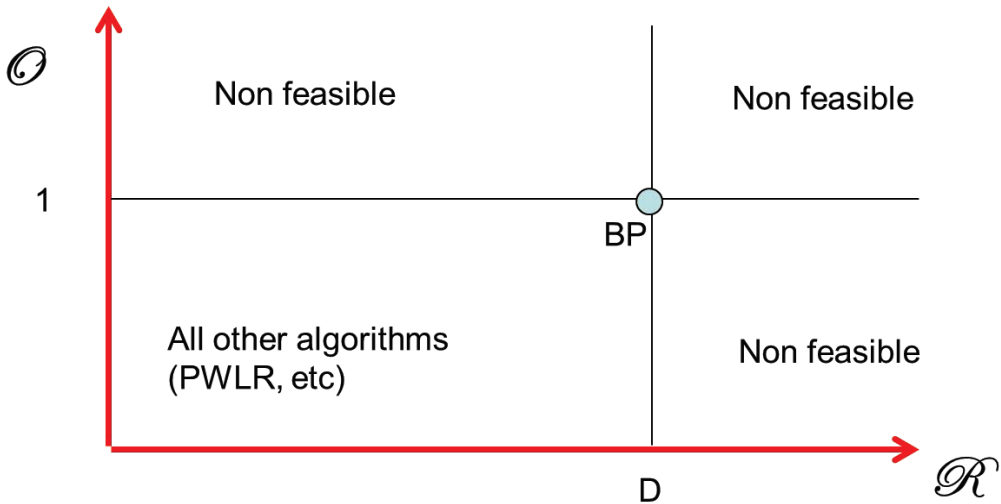}
\end{center}
\caption{\small Backpropagation is optimal in the space of possible learning algorithms, achieving both the maximal possible rate ${\cal R}=D$ and maximal expected improvement ${\cal O}=1$.}
\label{fig:algos}
\end{figure}

\subsection{Recurrent Networks}

Remarkably, the results of the previous sections can be extended to recurrent, as well as recursive, networks (see \cite{xie2003equivalence} for an attempt at implementing backpropagation in recurrent networks using Hebbian learning). To see this, consider a recurrent network with $W$ connection weights, where the connections can form directed cycles. If the network is unfolded in time over $L$ time steps, one obtains a deep feedforward network 
(Figure \ref{fig:RNN}), where the same sets of original weights from the recurrent networks is used to update all the unit activations, from one layer (or time step) to the next. Thus the unfolded version has a set of $W$ weights that are shared $L$ times. In the recurrent case, one may have targets for all the units at all the time steps, or more often, targets may be available only at some time steps, and possibly only for some of the units. Regardless of the pattern of available targets, the same argument used in Section 6.4 to expose the limitations of deep local learning, exposes the limitations of local learning in recurrent networks. More precisely, under the assumption that the error function is a differentiable function of the weights,
any algorithm capable of reaching an optimal set of weights-- where all the partial derivatives are zero--must be capable of ``backpropagating'' the information provided by any target at any time step to all the weights capable of influencing the corresponding activation. This is because a target $T_i^l$ for unit $i$ at time $l$ will appear in the partial derivative of any weight present in the recurrent network that is capable of influencing the activity of unity $i$ in $l$ steps or less. Thus, in general, an implementation capable of reaching an optimal set of weights must have a ``channel'' in the unfolded network capable of transmitting information from $T_i^L$ back to all the 
weights in all the layers up to $l$ that can influence the activity of unity $i$ in layer $l$. Again in a large recurrent network the maximal amount of information that can be sent back is the full gradient and the minimal number of operations required typically scales like $WL$. Thus this shows that the backpropagation through time algorithm is optimal in the sense of providing the most information, i.e. the full gradient, for the least number of computations $(WL)$.

Boltzmann machines \cite{ackley85}, which can be viewed as a particular class of recurrent networks with symmetric connections, can have hidden nodes and thus be considered deep. Although their main learning algorithm can be viewed as a form of simple Hebbian learning ($\Delta w_{ij} \propto <O_iO_j>_{clamped} - 
<O_iO_j>_{free})$, they are no exception to the previous analyses.
This is because the connections of a Boltzmann machines provide a channel allowing information about the targets obtained at the visible units to propagate back towards the deep units. Furthermore, it is well known that this learning rule precisely implements gradient descent with respect to the relative divergence between the true and observed distributions of the data, measured at the visible units. Thus the Hebbian learning rule for Boltzmann machines implements a form of local deep learning which in principle is capable of transmitting the maximal amount of information, from the visible units to the deep units, equal to the gradient of the error function. What is perhaps less clear is the computational cost and how it scales with the total number of weights $W$, since the learning rule in principle requires the achievement of equilibrium distributions.

Finally, the nature of the learning channel, and its temporal dynamics, in physical recurrent networks, including biological neural networks, are important but beyond the scope of this paper. However, the analysis provided is already useful in clarifying that the backward recurrent connections could serve at least three different roles: (1) a fast role to dynamically combine bottom-up and top-down activity, for instance during sensory processing; (2) a slower role to help carry signals for learning the feedforward connections; and (3) a slower role to help carry signals for learning the backward connections. 

\begin{figure}[h!]
\begin{center}
\includegraphics[width=0.8\columnwidth]{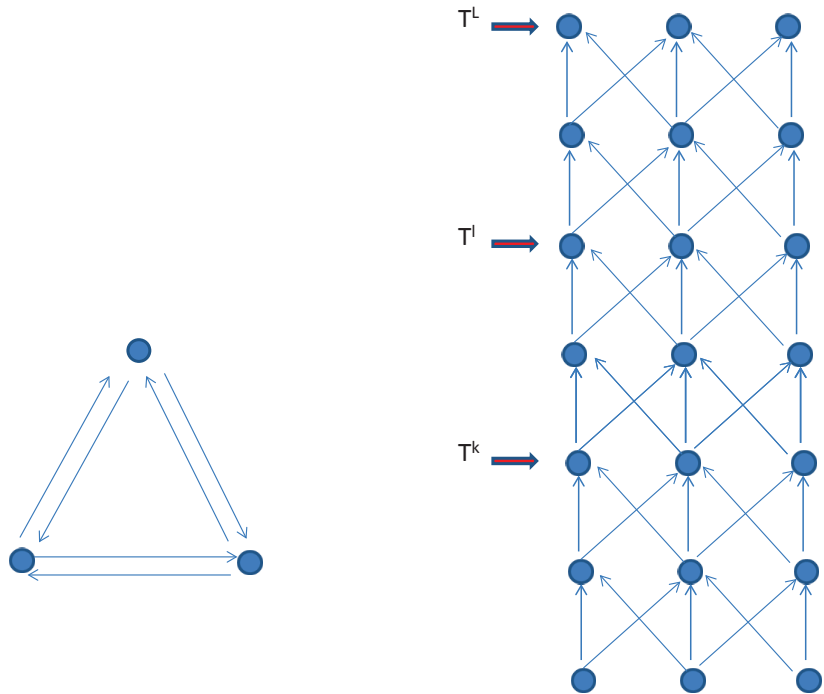}
\end{center}
\caption{\small Left: Recurrent neural network with three neurons and $W=6$ connection weights.  Right: The same network unfolded through time for $L$ steps, producing a deep feedforward network where the weights are shared between time steps. Each original weight is replicated $L$ times . Targets are available for a subset of the layers (i.e. time steps). In order to reach an optimal set of weights, a learning algorithm must allow each individual target to influence all the copies of all the weights leading to the corresponding unit.
}
\label{fig:RNN}
\end{figure}

\section{Conclusion}
The concept of Hebbian learning has played an important role in computational neuroscience, neural networks, and machine learning for over six decades. However, the vagueness of the concept has hampered systematic investigations and overall progress. To redress this situation, it is beneficial to expose two separate notions: the locality of learning rules and their functional form. Learning rules can be viewed as mathematical expressions for computing the adjustment of variables describing synapses during learning, as a function of variables which, in a physical system, must be 
local. Within this framework, we have studied the space of polynomial learning rules in linear and non-linear feedforward neural networks. In many cases, the behavior of these rules can be estimated analytically and reveals how these rules are capable of extracting relevant statistical information from the data. However, in general, deep local learning associated with the stacking of local learning rules in deep feedforward networks is not sufficient to learn complex input-output functions, even when targets are available for the top layer. 

Learning complex input-output functions requires a learning channel capable of propagating information about the targets to the deep weights and resulting in local deep learning. In a physical implementation, this learning channel can use either the forward connections in the reverse direction, or a separate set of connections. Furthermore, for large networks, all the information carried by the feedback channel can be interpreted in terms of the number of bits of information about the gradient provided to each weight. The capacity of the feedback channel can be defined in terms of the number of bits provided about the gradient per weight, divided by the number of required operations per weight. The capacity of many possible algorithms can be calculated, and the calculations show that backpropagation outperforms all other algorithms as it achieves the maximum possible capacity. This is true in both feedforward and recurrent networks.
It must be noted, however, that these results are obtained using somewhat rough estimates--up to multiplicative constants--and there may be other interesting algorithms that scale similarly to backpropagation. In particular, we are investigating the use of random, as opposed to symmetric, weights  in the learning channel, which seems to work in practice \cite{lillicrap2014random}.
  
The remarkable optimality of backpropagation suggests that when deploying learning systems in computer environments with specific constraints and budgets (in terms of big data, local storage, parallel implementations, communication bandwidth, etc) backpropagation provides the upper bound of what can be achieved, and the model that should be emulated or approximated by other systems.

Likewise, the optimality of backpropgation leads one to wonder also whether, by necessity, biological neural systems must have discovered some form of stochastic gradient descent during the course of evolution. While the question of whether the results presented here have some biological relevance is interesting, 
several other points must be taken into consideration.
First, the analyses have been carried in the simplified supervised learning setting, which is not meant to closely match how biological systems learn. Whether the supervised learning setting can approximate at least some essential aspects of biological learning is an open question, and so is the related question of extending the theory of local learning to other forms of learning, such as reinforcement learning \cite{sutton1998reinforcement}. 

Second, the analyses have been carried using artificial neural network models. Again the question of whether these networks capture some essential properties of biological networks is not settled. Obviously biological neurons are very complex biophysical information processing machines, far more complex than the neurons used here. On the other hand, there are several examples in the literature (see, for instance, \cite{zipser1988back,olshausen1996emergence,
polsky2004computational,poggio14,yamins2016using}) where important biological properties seem to be captured by artificial neural network models. [In fact these results, taken together with the sometimes superhuman performance of backpropagation and the optimality results presented here, lead us to conjecture paradoxically 
that biological neurons may be trying to approximate artificial neurons, and not the other way around, as has been assumed for decades.] But even if they were substantially unrelated to biology, artificial neural networks still provide the best simple model we have of a connectionist style of computation and information storage, entirely different from the style of digital computers, where information is both scattered and superimposed across synapses and intertwined with processing, rather than stored at specific memory addresses and segregated from  processing. 

In any case, for realistic biological modeling, the complex geometry of neurons and their dendritic trees must be taken into consideration. For instance, there is a significant gap between 
having a feedback error signal $B_i$ arrive at
the soma of neuron $i$, and having $B_i$ available as a local variable in a far away synapse located in the dentritic tree of neuron $i$. In other words, $B_i$ must become a local variable at the synapse $w_{ij}$. Using the same factor of $10^6$ from the Introduction, which rescales a synapse to the size of a fist, this gap could correspond to tens or even hundreds of meters. Furthermore, in a biological or other physical system, one must worry about locality not only in space, but also in {\it time}, e.g. how close must 
$B_i$ and $O_j$ be in time?

Third, issues of coordination of learning across different brain 
components and regions must also be taken into consideration (e.g. \cite{valiant2012hippocampus}).
And finally, a more complete model of biological learning would have to include  not only target signals that are backpropagated electrically, but ultimately also the complex and slower biochemical processes involved in synaptic modification, including gene expression and epigenetic modifications, and the complex production, transport, sequestration, and degradation of protein, RNA, and other molecular species (e.g. \cite{guan2002integration,
mayford2012synapses,woodnatureneuroscience12}).

However, while there is no definitive evidence in favor or against the use of stochastic gradient descent in biological neural systems, and obtaining such evidence remains a challenge, biological deep learning must follow the locality principle and thus the theory of local learning provides a framework for investigating this fundamental question.

\section*{Appendix A: Uniqueness of Simple Hebb in Hopfield  Networks}

A Hopfield model can be viewed as a network of $N$ $[-1,1]$ threshold gates connected symmetrically ($w_{ij}=w_{ji}$) with no self-connections ($w_{ii}=0$). As a result the network has a quadratic energy function 
$E=-(1/2)\sum_{ij} w_{ij} O_iO_j$ and the dynamics of the network under stochastic asynchronous updates converges to local minima of the energy function.
Given a set $\cal S$ of $M$ memory vectors $M^{(1)},M^{(2)}, \ldots M^{(M)}$, the simple Hebb rule is used to produce an energy function $E_{\cal S}$ to try to store these memories as local minima of the energy function so that
$w_{ij}=\sum_k M_i^{(k)}M_j^{(k)}$. $E_{\cal S}$ induces an acyclic orientation $\cal O (S)$ of the $N$ dimensional hypercube $\cal H$. If $h$ is an isometry of $\cal H$ for the Hamming distance, then for the simple Hebb rule we have ${\cal O}(h({\cal S}))=h({\cal O}({\cal S}))$. Are there any other learning rules with the same property?

We consider here learning rules with $d=0$. Thus we must have 
$\Delta w_{ij}=F(O_i,O_j)$ where $F$ is a polynomial function. On the [-1,1] hypercube, we have $O_i^2=O_j^2=1$ and thus we only need to consider the case $n=2$ with $F(O_i,O_j)=\alpha O_iO_j + \beta O_i +\gamma O_j + \delta$. However the learning rule must be symmetric in $i$ and $j$ to preserve the symmetry $w_{ij}=w_{ji}$. Therefore $F$ can only have the form
$F(O_i,O_j)=\alpha O_iO_j + \beta (O_i + O_j) +\gamma$. 
Finally, the isometric invariance must be true for {\it any} set of memories $\cal S$. It is easy to construct examples, with specific sets $\cal S$, that force $\beta$ and $\gamma$ to be 0. Thus in this sense the simple Hebb rule 
$\Delta w_{ij}=\alpha O_iO_j$ is the only isometric invariant learning rule for the Hopfield model. A similar results can be derived for spin models with higher-order interactions where the energy function is a polynomial of degree $n>2$ in the spin variables \cite{baldi88a}.

\section*{Appendix B: Invariance of the Gradient Descent Rule}
In the $[0,1]$ case with the logistic transfer function,
the goal is to minimize the relative entropy error

\be
E=- \left [ T \log O + (1-T) \log (1-O) \right ]
\label{eq:}
\ee
Therefore

\be
\frac{\partial E}{\partial O} =\frac{T-O}{O(1-O)}
\quad {\rm and } \quad \frac{\partial O}{\partial S}=O(1-O)
\quad {\rm and} \;\; {\rm thus} \quad
\Delta w_i =\eta (T-O)I_i
\label{eq:}
\ee
In the $[-1,1]$ case with the $\tanh$ transfer function, 
the equivalent goal is to minimize

\be
E'=- \left [ \frac{T'+1}{2}\log \frac {O'+1}{2}
+ \frac{1-T'}{2} \log \frac {1-O'}{2} \right ]
\label{eq:}
\ee
where $T=\frac{T'1+1}{2}$ and $O=\frac{O'+1}{2}$.
Therefore

\be
\frac{\partial E'}{\partial O'} =\frac{2(T'-O')}{1-{O'}^2}
\quad {\rm and } \quad \frac{\partial O'}{\partial S'}=1-{O'}^2
\quad {\rm and} \;\; {\rm thus} \quad
\Delta w'_i =\eta 2(T'-O')I'_i
\label{eq:}
\ee
Thus the gradient descent learning rule is the same, up to a factor of 2 which can be absorbed by the learning rule. 
The origin of this factor lies in the fact that $\tanh (x) =
(1-e^{-2x})/(1+e^{-2x})$ is actually {\it not} the natural $[-1,1]$ equivalent of the logistic function $\sigma (x)$. The natural equivalent is 

\be
\tanh \frac{x}{2} =2\sigma(x)-1=\frac{1-e^{-x}}{1+e^{-x}}
\label{eq:}
\ee

\section*{Appendix C: List of New Convergent Learning Rules}

All the rules are based on adding a simple decay term to the simple Hebb rule and its supervised variants.

\subsection*{Fixed Decay}

\be 
\Delta w_{ij} \propto O_iO_j - Cw_{ij} \quad {\rm with} \quad C>0
\label{eq:nr10}
\ee
with the supervised clamped version

\be 
\Delta w_{ij} \propto T_iO_j - C w_{ij}
\label{eq:nr1}
\ee
and the gradient descent version

\be 
\Delta w_{ij} \propto (T_i-O_i)O_j - C w_{ij}
\label{eq:nr12}
\ee

\subsection*{Adaptive Decay Depending on the Presynaptic Term}

\be 
\Delta w_{ij} \propto O_iO_j - O_j^2 w_{ij}
\label{eq:nr13}
\ee
with the supervised clamped version

\be 
\Delta w_{ij} \propto T_iO_j - O_j^2 w_{ij} 
\quad 
\label{eq:nr14}
\ee
and the gradient descent version

\be 
\Delta w_{ij} \propto (T_i-O_i)O_j - O_j^2 w_{ij}
\label{eq:nr15}
\ee

\subsection*{Adaptive Decay Depending on the Postsynaptic Term}

\be 
\Delta w_{ij} \propto O_iO_j - O_i^2 w_{ij}
\label{eq:nr16}
\ee
This is Oja's rule, which yields the supervised clamped versions

\be 
\Delta w_{ij} \propto T_iO_j - O_i^2 w_{ij} \quad
{\rm and} \quad \Delta w_{ij} \propto T_iO_j - T_i^2 w_{ij}
\label{eq:nr17}
\ee
and the gradient descent versions

\be 
\Delta w_{ij} \propto (T_i-O_i)O_j - O_i^2 w_{ij} \quad 
{\rm and} \Delta w_{ij} \propto (T_i-O_i)O_j - (T_i-O_i)^2 w_{ij}  
\label{eq:nr18}
\ee

\subsection*{Adaptive Decay Depending on the Pre- and Post-Synaptic (Simple Hebb) Terms}

\be 
\Delta w_{ij} \propto O_iO_j - (O_iO_j)^2 w_{ij}=O_iO_j(1-O_iO_jw_{ij})
\label{eq:nr19}
\ee
with the clamped versions

\be 
\Delta w_{ij} \propto T_iO_j - (O_iO_j)^2 w_{ij}\quad
{\rm and} \Delta w_{ij} \propto T_iO_j - (T_iO_j)^2 w_{ij}
\label{eq:nr20}
\ee
and gradient descent versions

\be 
\Delta w_{ij} \propto (T_i- O_i)O_j - (O_iO_j)^2 w_{ij} \quad
{\rm and} \Delta w_{ij} \propto (T_i- O_i)O_j - ((T_i-O_i)O_j)^2 w_{ij} 
\label{eq:nr21}
\ee

We now consider the alternative approach which bounds the weights in a $[-C,C]$ range for some $C<0$. The initial values of the weights are assumed to be small or 0.

\subsection*{Bounded Weights}

\be 
\Delta w_{ij} \propto O_iO_j (C-w_{ij}^2)
\label{eq:nr22}
\ee
with the clamped version

\be 
\Delta w_{ij} \propto T_iO_j (C-w_{ij}^2)
\label{eq:nr23}
\ee
and the gradient descent version

\be 
\Delta w_{ij} \propto (T_i- O_i)O_j (C-w_{ij}^2)
\label{eq:nr24}
\ee

\section*{Appendix D: Additional Remarks on Deep Targets Algorithms}

\noindent
{\bf 1)} In many situations, for a given input vector $I$ there will be a corresponding and distinct activity vector $O^{h-1}$. However, sometimes the function $C_{h-1}$ may not be injective in which cases several input vectors $I(t_1), \ldots ,I(t_k)$, with final targets $T(t_1), \ldots ,T(t_k)$, may get mapped onto the same activity vector $O^{h-1}=C_{h-1}(I(t_1))= \ldots = C_{h-1}(I(t_k))$. In this case, the procedure for determining the target vector $T^h$ may need to be adjusted slightly as follows. First, the sample of activity ${\cal S}^h$ is generated and propagated forward using the function $A_{h+1}$, as in the non-injective case. However the selection of the best output vector over the sample may take into consideration all the targets $T(t_1), \ldots ,T(t_k)$ rather than the isolated target associated with the current input example. For instance, the best output vector may be chosen as to minimize the sum of the errors with respect to all these targets.
This procedure is the generalization of the procedure used to train an unrestricted Boolean autoencoder \cite{baldiboolean12}.
\par
\noindent
{\bf 2)} Depending on the schedule in the outer loop, the sampling approach, and the optimization algorithm used in the inner loop, as well as other implementation details, the description above provides a family of algorithms, rather than a single algorithm.
Examples of schedules for the outerloop include a single pass from layer 1 to layer $L$, alternating up-and down passes along the architecture,  cycling through the layers in the order 1,2,1,2,3,1,2,3,4, etc, and their variations.
\par
\noindent
{\bf 3)} The sampling deep targets approach can be combined with all the other ``tricks'' of backpropagation such as weight sharing and convolutional architectures, momentum, dropout, 
and so forth. Adjustable learning rates can be used with different adjustment rules for different learning phases \cite{bottou-98x,bottou-mlss-2004}. 
\par
\noindent
{\bf 4)} The sampling deep targets approach can be easily combined also with backpropagation.
For instance, targets can be provided for every other layer, rather than for every layer, and backpropagation used to train pairs of adjacent layers. It is also possible to interleave the layers over which backpropagations is applied to better stitch the shallow components together (e.g. use backpropagations for layers 3,2,1 then 4,3,2, etc).
\par
\noindent
{\bf 5)} When sampling from a layer, here
we have focused on using the optimal output sample to derive the target.
It may be possible instead to leverage additional information
contained in the entire distribution of  samples. 
\par
\noindent
{\bf 6)} In practice the algorithm converges, at least to a local minima of the error function. In general the convergence is not monotonic
(Figure \ref{fig:thresholdgate_auto}), with occasional uphill jumps that can 
be beneficial in avoiding poor local minima. Convergence can be proved mathematically in several cases. For instance,
if the optimization procedure can map each hidden activity to each corresponding target over the entire training set, then the overall training error is guaranteed  to decrease or stay constant at each optimization step and hence it will converge to a stable value. In the unrestricted Boolean case (or in the Boolean case with perfect optimization), with exhaustive sampling of each hidden layer the algorithm can also be shown to be convergent. 
Finally, it can also be shown to be convergent in the framework of stochastic learning and stochastic component optimization
\cite{robbins1971convergence,bottou-mlss-2004}.
\par
\noindent
{\bf 7} A different kind of deep targets algorithm, where the output targets are used as targets for all the hidden layers, is described in \cite{deepcontact2012}. The goal in this case is to force successive hidden layers to refine their predictions towards the final target.

\paragraph{Acknowledgments}
Work supported in part by NSF grant 
IIS-1550705 and a Google Faculty Research Award to PB.
We are also grateful for a hardware gift from NVDIA Corporation. This work was presented as a keynote talk at the 2015 ICLR Conference and a preliminary version was posted on ArXiv under the title
``The Ebb and Flow of Deep Learning''.

\bibliography{baldi,nn,info,math}

\bibliographystyle{abbrv}
\end{document}